\pdfoutput=1

\PassOptionsToPackage{table}{xcolor}

\documentclass[11pt]{youtu} 

\usepackage{mathpazo}

\usepackage[T1]{fontenc}
\usepackage[utf8]{inputenc}

\usepackage{natbib}
\setcitestyle{numbers,square,comma}

\usepackage{graphicx}
\usepackage{adjustbox}

\usepackage{latexsym}
\usepackage{booktabs}
\usepackage{multicol}
\usepackage{multirow, makecell}
\usepackage{colortbl}
\usepackage{tikz}
\usepackage{arydshln}
\usepackage{pgfplots}
\pgfplotsset{compat=1.18}
\usepackage{amsmath}
\usepackage{float}
\usepackage{amssymb}
\usepackage{marvosym}
\usepackage{tabularx}
\usepackage{enumerate}
\usepackage{enumitem}
\usepackage{cuted}
\definecolor{softblue}{RGB}{232,244,255}
\definecolor{softgreen}{RGB}{0,150,0}

\usepackage{algorithm}
\usepackage{algpseudocode}
\usepackage{subcaption}

\definecolor{darkgreen}{RGB}{0, 100, 0}
\definecolor{softred}{RGB}{255,235,235}
\definecolor{softorange}{RGB}{255,242,224}

\usepackage{listings}

\title{VideoSearcher: Empowering Video Deep Research with Multi-Tool Agentic Reasoning via Reinforcement Learning}

\renewcommand{\authorlist}{%
  \begin{center}
    {\normalsize
      Zhenkun Gao$^{*,1,2}$, Yicheng Bao$^{*,1}$, Jinlong Peng$^{*,\dagger,\text{\Letter},2}$, Xueheng Li$^{*,\text{\Letter},3}$, Theo Huang$^{*,4,2}$\\
      Bangwei Liu$^{1}$, Kunquan Li$^{5}$, Zhenye Gan$^{2}$, Tao Hu$^{3}$, Chengjun Xie$^{3}$, Mingqian Yang$^{1,2}$\\
      Xuanhua He$^{\text{\Letter},6}$, Zhizhong Zhang$^{1}$, Xin Tan$^{1}$, Chengjie Wang$^{2}$, Yuan Xie$^{\text{\Letter},1}$%
    }
  \end{center}
}
\renewcommand{\affiliationlist}{%
  \begin{center}
    {\footnotesize
      $^{*}$ Equal Contribution,\quad $^{\dagger}$ Project Leader,\quad \Letter{} Corresponding Author\\[0.25em]
      $^{1}$East China Normal University,\quad $^{2}$Tencent Youtu Lab,\quad $^{3}$HFIPS, Chinese Academy of Sciences\\
      $^{4}$Wuhan University,\quad $^{5}$Xiamen University,\quad $^{6}$The Hong Kong University of Science and Technology%
    }
  \end{center}
}

\sourcecode{https://github.com/Stephen-gzk/VideoSearcher}
\model{https://huggingface.co/Stephengzk/VideoSearcher-8B}
\dataset{https://huggingface.co/datasets/Stephengzk/VideoSearcher-Data}
\project{https://stephen-gzk.github.io/VideoSearcher-website/}

\begin{document}


\abstract{
Video understanding is moving beyond closed-context perception toward open-world evidence exploration, a paradigm formalized as Video Deep Research (VDR). However, existing multimodal search agents primarily target static images, and the current VDR benchmark relies on text-centric retrieval that discards crucial visual information. To address these limitations, we propose VideoSearcher, a closed-loop agentic framework that empowers Vision-Language Models with multi-tool reasoning for VDR. VideoSearcher unifies temporal localization, spatial focusing, and multimodal search within a single reasoning trajectory, enabling agents to progressively ground visual clues, retrieve relevant evidence, and synthesize answers.  To optimize knowledge-intensive reasoning trajectories, we propose Bi-branch Sequence Policy Optimization (BiSPO), a reinforcement learning algorithm that decouples tool-invocation optimization from answer-accuracy optimization. This design provides stable learning signals for both evidence-grounded reasoning and purposeful tool use. Furthermore, we construct VideoSearch-QA, the first benchmark designed to evaluate open-world video information grounding and multimodal search-based reasoning. Extensive experiments demonstrate that VideoSearcher significantly outperforms prior open-source agentic baselines across various search-oriented and multimodal understanding benchmarks.
}

\maketitle

\renewcommand{\thefootnote}{\textdagger}

\section{Introduction}
Video understanding \cite{team2025vidi, fu2025video, ning2025video} remains a critical bottleneck for deploying Vision-Language Models (VLMs) \cite{bai2025qwen3, wang2025internvl3} in real-world applications \cite{zhang2025thinking, feng2026video}. Unlike static images, videos encompass evolving temporal dynamics, requiring models to track state transitions across frames for accurate spatio-temporal modeling. Particularly in long videos \cite{chen2026scaling}, crucial evidence is frequently submerged within massive redundant frames and manifests only during fleeting temporal windows. Therefore, models need to localize the relevant moments in time, focus on fine-grained visual details such as text or small objects, and reason over the collected evidence to answer the query. Robust video understanding thus requires both global context modeling and fine-grained spatio-temporal perception~\cite{tang2025video}.

\begin{figure}[t]
  \centering
  \includegraphics[width=0.88\textwidth]{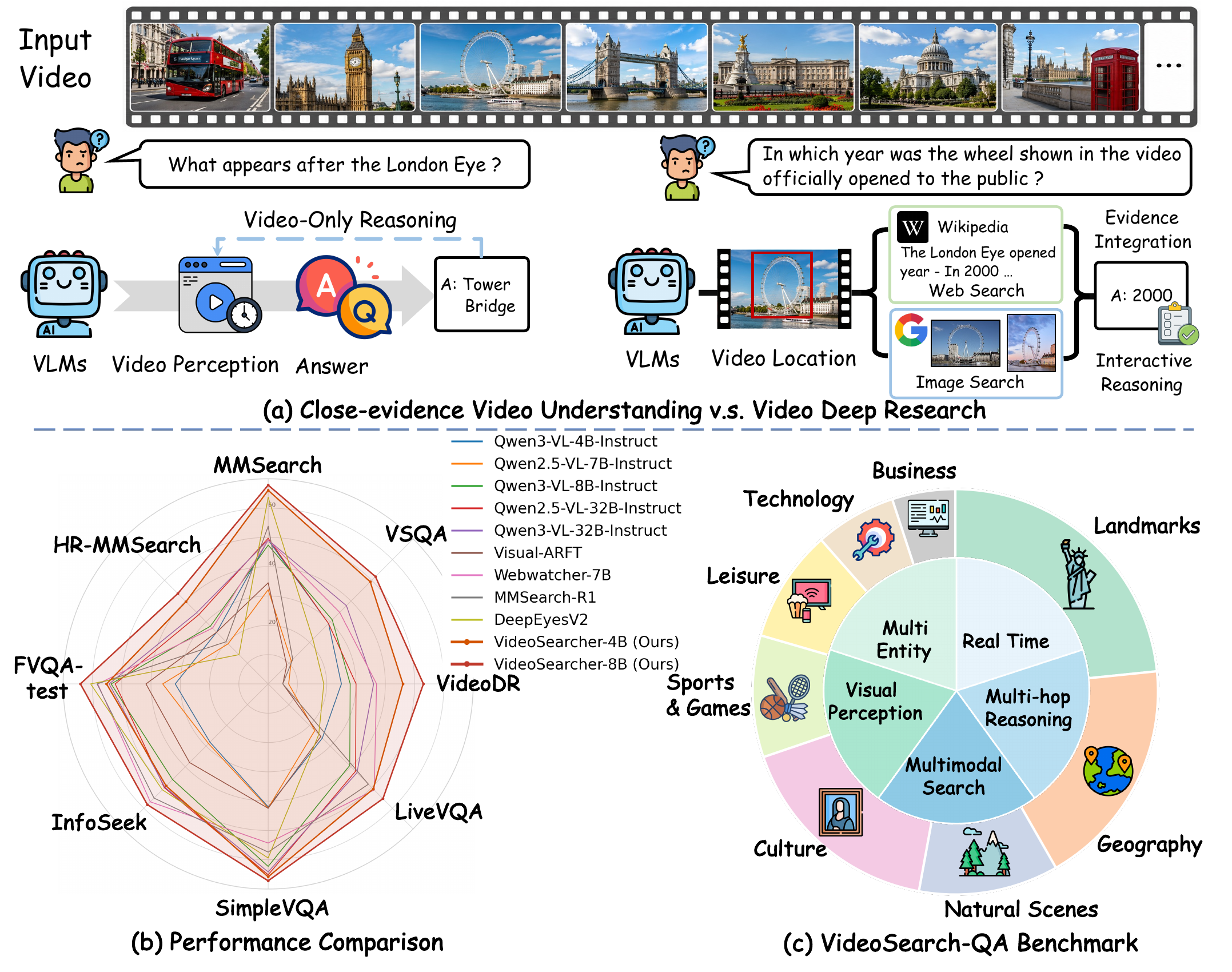}
  \caption{Overview of {VideoSearcher}. (a) Video Deep Research extends closed-evidence video understanding by requiring agents to ground visual cues and retrieve open-web evidence. (b) Performance comparison on search-oriented benchmarks. (c) Our constructed {VideoSearch-QA} benchmark covers diverse domains for evaluating video-grounded multimodal search ability.}
  \label{fig:teaser}
\end{figure}

However, most existing video understanding works \cite{he2025framethinker, yang2025longvt} adhere to a closed-evidence paradigm, assuming that all necessary information resides within the input video. While this setting is useful for evaluating spatio-temporal video understanding, it remains insufficient for real-world scenarios. In practical video QA, videos often provide only visual anchors (e.g., logos or people), while the final answer must be retrieved from open-web sources, including webpages, images, news, or knowledge bases, as shown in Fig.~\ref{fig:teaser}(a). Recently, VideoDR~\cite{liu2026watching} proposed a benchmark and formalized this setting as \textbf{Video Deep Research} (VDR), where models must identify cross-frame visual anchors and combine them with open-web retrieval to produce verifiable answers. This paradigm extends video understanding from closed-context perception to open-world evidence exploration, positioning videos as entry points for agents to retrieve and verify information in expansive web environments.

\begin{figure}[t]
  \centering
  \includegraphics[height=0.50\textheight,keepaspectratio]{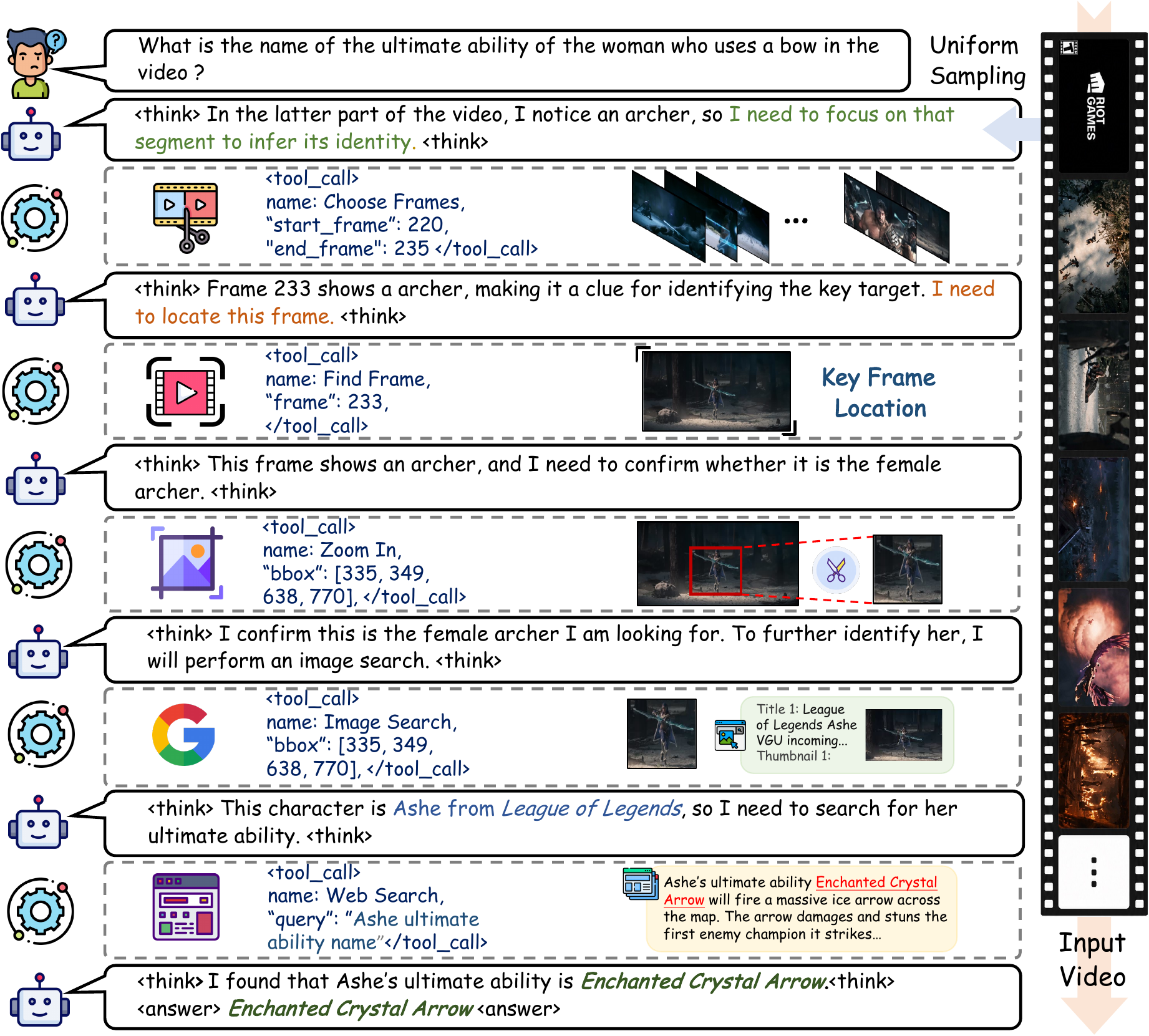}
  \caption{Agentic reasoning trajectory of {VideoSearcher}. {VideoSearcher} performs agentic reasoning by localizing key frames, inspecting regions, invoking web/image search, and integrating multimodal evidence for answer generation.}
  \label{fig:reasoning_trajectory}
\end{figure}

In recent years, deep research \cite{jin2025search} and multimodal search \cite{narayan2025deepmmsearch} agents have shown that tool-augmented closed-loop reasoning can substantially improve knowledge-intensive problem solving. Text-based agents solve open-web QA through iterative query formulation and evidence integration \cite{li2025search}, while multimodal agents \cite{wu2025mmsearch, hong2025deepeyesv2} further introduce visual tools such as image cropping and image search to combine external knowledge with localized visual analysis. However, these methods are primarily designed for static images, making them insufficient for VDR, where agents must track cross-frame cues, model spatio-temporal state changes, and jointly decide temporal localization, spatial focus, and search modality. Although VideoDR \cite{liu2026watching} represents an important step, it remains largely text-centric by converting visual cues into textual descriptions before web search. This setting loses the original visual information and cannot fully reflect  real-world video retrieval. Capable VDR agents should localize visual anchors across frames, inspect fine-grained regions, construct multimodal queries, and reason over evidence gathered through long-horizon web interaction.  This demands fine-grained video understanding and precise tool use, including discriminative text queries and targeted cropped regions for image search.

To address these challenges, we introduce \textbf{VideoSearcher}, a pioneering agentic reasoning and search framework for Video Deep Research. VideoSearcher equips the model with five tools: \texttt{choose\_frames} for coarse temporal selection, \texttt{find\_frame} for key frame localization, \texttt{zoom\_in} for local region inspection, as well as \texttt{web\_search} and \texttt{image\_search} for textual and visual web retrieval. These tools enable a unified trajectory in which the model progressively narrows the video timeline, focuses on informative regions, retrieves multimodal evidence, and integrates the acquired evidence for reasoning. We adopt a two-stage training paradigm comprising cold-start Supervised Fine-tuning (SFT) and Reinforcement Learning (RL). To construct multi-tool reasoning trajectories, we design a VDR-specific data synthesis pipeline. We extract and cross-validate core entities from search-oriented image-text QA datasets, and use them as retrieval targets to collect relevant videos. The retrieved videos are normalized into unified frame sequences. We then rewrite the original QA pairs to align with the video context, followed by visual-QA alignment verification to discard noisy samples with missing entities, answer leakage, or inconsistent visual anchors. We then synthesize trajectories by first localizing key frames and  generating complete reasoning traces with diverse tool invocations. Through rigorous filtering and quality assurance, we derive high-quality SFT data. The remaining hard video-QA instances are reserved for RL rollouts in the VDR environment.


Furthermore, we propose Bi-branch Sequence Policy Optimization (\textbf{BiSPO}) to optimize long-horizon multi-tool reasoning trajectories in VDR. Built upon GSPO \cite{zheng2025group}, BiSPO applies sequence-level importance sampling and clipping, and further separates answer accuracy optimization from tool invocation optimization. This dual-branch design provides more stable learning signals for knowledge-intensive and reasoning-heavy VDR. We also introduce a bell-shaped gated tool invocation reward that encourages sufficient video observation and external retrieval within a reasonable range, while penalizing excessive tool calls that may cause invalid searches or incorrect tool sequences. Finally, we introduce \textbf{VideoSearch-QA} (VSQA), a benchmark for evaluating whether VLM-based agents can jointly capture video cues and perform multimodal search-based reasoning in dynamic videos. Extensive experiments on VideoDR \cite{liu2026watching}, VideoSearch-QA, and a suite of multimodal search and video understanding benchmarks validate the effectiveness of VideoSearcher.

In summary, our contributions are as follows:
\begin{itemize}[leftmargin=*, itemsep=1pt, topsep=1pt, parsep=1pt, partopsep=1pt]

    \item We propose VideoSearcher, the first closed-loop Video Deep Research agent that integrates temporal localization, spatial focusing and multimodal search for dynamic videos.
    \item We propose {BiSPO}, a dual-branch RL algorithm that separately optimizes answer accuracy and tool invocation behavior for stable long-horizon tool-intensive reasoning.
    \item We construct {VideoSearch-QA}, the first benchmark for evaluating VDR agents on video information grounding and multimodal search-based reasoning in open-world scenarios. 
    \item Extensive experiments on VideoDR, VSQA, and multiple multimodal search and video understanding benchmarks validate the effectiveness of VideoSearcher.
\end{itemize}


\section{Related Work}
\subsection{Video Understanding}
Recent advances in Vision-Language Models (VLMs) have substantially improved video understanding, particularly in spatio-temporal perception and reasoning \cite{bai2025qwen3, gao2026tpru}. Advanced video-centric models such as Video-R1 \cite{feng2026video} and VideoRFT \cite{wang2026videorft}  leverage Reinforcement Learning to strengthen the video reasoning capabilities of models. FrameThinker \cite{he2025framethinker} and LongVT \cite{yang2025longvt} further introduce agentic techniques, such as iterative tool use and adaptive frame selection, to improve long video understanding. Despite these advances, most existing works still follow a closed-evidence paradigm \cite{liu2026watching}, where answers are assumed to be inferable solely from the input video, leaving open-web information acquisition and multimodal search-oriented reasoning insufficiently explored.

\subsection{Tool-Augmented Agentic VLMs}
Existing studies have advanced VLMs from passive perception to tool-augmented agentic reasoning.  Under the ``thinking with images'' paradigm, models such as OpenThinkIMG~\cite{su2025openthinkimg} and DeepEyesV2~\cite{hong2025deepeyesv2} use interactive visual tools to manipulate, generate or inspect intermediate visual states for fine-grained reasoning.  In addition, search agents such as MMSearch-R1 \cite{wu2025mmsearch} and Webwatcher \cite{geng2025webwatcher} integrate multimodal search tools, enabling models to acquire open-web information beyond their parametric knowledge. Other efforts \cite{qiao2025v, chu2026redsearcher} further explore tool use for complex visual reasoning, including visual analysis and code execution. Despite these advances, existing models are mainly designed for text or static images. They remain insufficient for VDR, where agents must perform cross-frame localization, spatial focusing, multimodal search, and long-chain evidence reasoning over dynamic videos. 
 

\section{Methods}
\subsection{Problem Formulation} 
\noindent \textbf{Task Definition.} We formulate Video Deep Research (VDR) as an open-world video question answering task. Given a video \(\mathcal{V}\), a  question \(q\), and an external web evidence space \(\mathcal{W}\), an agent \(\pi_\theta\) is required to produce an answer \(y\) through a multi-step reasoning trajectory \(\tau=(a_1,o_1,\ldots,a_T,o_T)\). At each turn \(t\), the agent selects an action \(a_t\in\mathcal{A}\), such as temporal localization, spatial inspection, image search, or web search, and receives an observation \(o_t\) from either the video or the open web. The trajectory induces a joint evidence set \(\mathcal{E}_\tau=\mathcal{E}^v_\tau\cup\mathcal{E}^w_\tau\), where \(\mathcal{E}^v_\tau\) contains localized video evidence such as frames, regions, entities, and visual texts, and \(\mathcal{E}^w_\tau\subseteq\mathcal{W}\) contains retrieved webpages, images, or textual snippets. Therefore, VDR requires agents to link sparse cross-frame anchors with multimodal search and reason over video and web evidence.


\noindent \textbf{Observation Space.} At turn \(t\), the agent observes the interaction history $\mathcal{H}_t$, where each \(o_i\in\mathcal{O}\) is the compact multimodal output returned by the corresponding tool call. The observation space is defined as \(\mathcal{O}=\mathcal{O}^{v}\cup\mathcal{O}^{w}\), where \(\mathcal{O}^{v}\) contains video-grounded observations (e.g., selected frames or localized regions) and \(\mathcal{O}^{w}\) contains web-grounded observations  (e.g., retrieved images or webpages). After executing \(a_t\), the returned observation \(o_t\) is appended to the history to form \(\mathcal{H}_{t+1}=(\mathcal{H}_t,a_t,o_t)\).

\noindent \textbf{Action Space.} At each turn, the agent first generates a reasoning state and then executes one valid action from \(\mathcal{A}\):
\begin{enumerate}[leftmargin=*, itemsep=0pt, topsep=2pt, parsep=0pt, partopsep=0pt]
    \item \texttt{choose\_frames}: Select a coarse temporal interval from the video.
    \item \texttt{find\_frame}: Localize a frame from the video.
    \item \texttt{zoom\_in}: Inspect a specified region of a frame.
    \item \texttt{web\_search}: Retrieve web evidence using a textual query.   
    \item \texttt{image\_search}: Perform visual retrieval with a frame or cropped region.
    \item \texttt{answer}: Produce the final prediction.
\end{enumerate}
It is noted that validity constraints require spatial and image-search actions to be grounded in selected frames with meaningful bounding boxes, and each step to include both a reasoning state and a well-formed action. Non-terminal actions return observations that update the interaction history.

\begin{figure}[t]
    \centering
    \includegraphics[width=\linewidth]{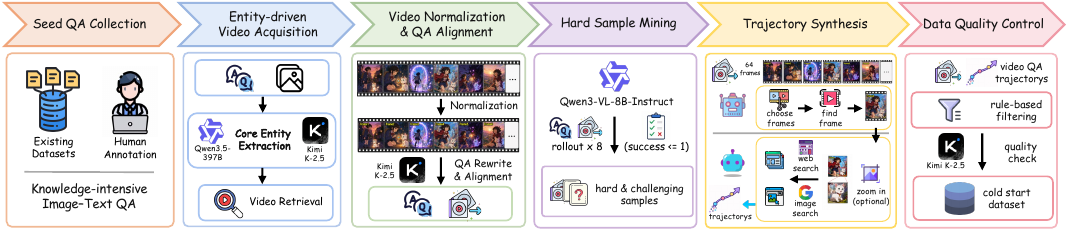}
    \caption{The video-centric training data synthesis pipeline of VideoSearcher.}
    \label{fig:data_pipeline}
\end{figure}
\subsection{Data Construction}
\subsubsection{Video-Centric Data Pipeline} 
To equip VideoSearcher with video grounding and tool-augmented deep research abilities, we build an automated video-centric data pipeline, as shown in Fig.~\ref{fig:data_pipeline}. The pipeline converts knowledge-intensive image-text QA samples into VDR instances, synthesizes multi-tool reasoning trajectories, and applies strict quality control before training. Detailed   procedures are provided in Appendix~\ref{appendix:a.1}.

\noindent \textbf{Entity-driven Video Acquisition.}  We start from knowledge-intensive image-text QA datasets, including FVQA~\cite{wu2025mmsearch}, DeepEyes~\cite{zheng2025deepeyes} and DeepEyesV2~\cite{hong2025deepeyesv2}. For each QA pair, we extract the core visual entities required for answering (e.g., landmarks, logos and people). To improve reliability, we query Qwen3.5-397B~\cite{qwen3.5} and Kimi-K2.5~\cite{team2026kimi}, and retain only entities consistently identified by both models. These verified entities are then used as retrieval targets to collect relevant videos from online platforms.

\noindent \textbf{Video Normalization and QA Alignment.} We normalize collected videos by resampling them to 1 FPS and overlaying a ``Frame \(N\)'' marker as a temporal reference. We then rewrite the original image-text QA pairs to align with the video context. The visual-QA alignment verification is applied to discard noisy samples with missing key entities, answer leakage, direct OCR exposure of the answer, or inconsistent visual anchors.

%



\noindent \textbf{Hard Sample Mining.} To focus on challenging VDR instances, we evaluate each candidate with Qwen3-VL-8B-Instruct and retain samples that the model consistently fails to solve. The resulting hard samples emphasize temporal localization, fine-grained visual inspection, and external evidence acquisition. We reserve \(3{,}285\) video instances for RL online rollouts, while the remaining samples are used for SFT trajectory synthesis.


\noindent \textbf{Hierarchical Trajectory Synthesis.} For each retained sample, we synthesize multi-tool trajectories with two proprietary models. A lightweight closed-source model performs coarse temporal screening via \texttt{choose\_frames}, while a more capable closed-source model handles fine-grained localization, multimodal web retrieval, and evidence reasoning within a limited turn budget. 



\noindent \textbf{Data Quality Control.} Strict quality control is applied before assembling the final SFT dataset. Heuristic filters remove trajectories with invalid tool invocations, malformed bounding boxes, and repetitive reasoning. Samples with hallucinated evidence or broken reasoning chains are further rejected by Kimi-K2.5. This process yields \(3{,}811\) high-quality SFT trajectories, which provide VideoSearcher with diverse tool-use patterns and basic VDR interaction capabilities.


\subsubsection{VideoSearch-QA Construction} 
VideoDR~\cite{liu2026watching} first formalizes VDR, but its evaluation remains largely text-centric, as visual cues are transformed into textual descriptions before web search. We introduce {VideoSearch-QA} (VSQA) as shown in Fig. \ref{fig:teaser}(c) and Fig. \ref{fig:vsqa_pipeline}, a manually annotated benchmark of \(278\) samples for evaluating multimodal video-grounded search, covering diverse domains such as landmarks, geography, culture, and recent 2026 events. For each video, we manually annotate fine-grained, search-oriented questions around key visual cues, enabling the evaluation of whether agents can capture visual information in dynamic videos and use it for open-web search and evidence-based reasoning. More details can be found in the Appendix \ref{appendix:a.2}.


\begin{figure}[t]
  \centering
  \includegraphics[width=\textwidth]{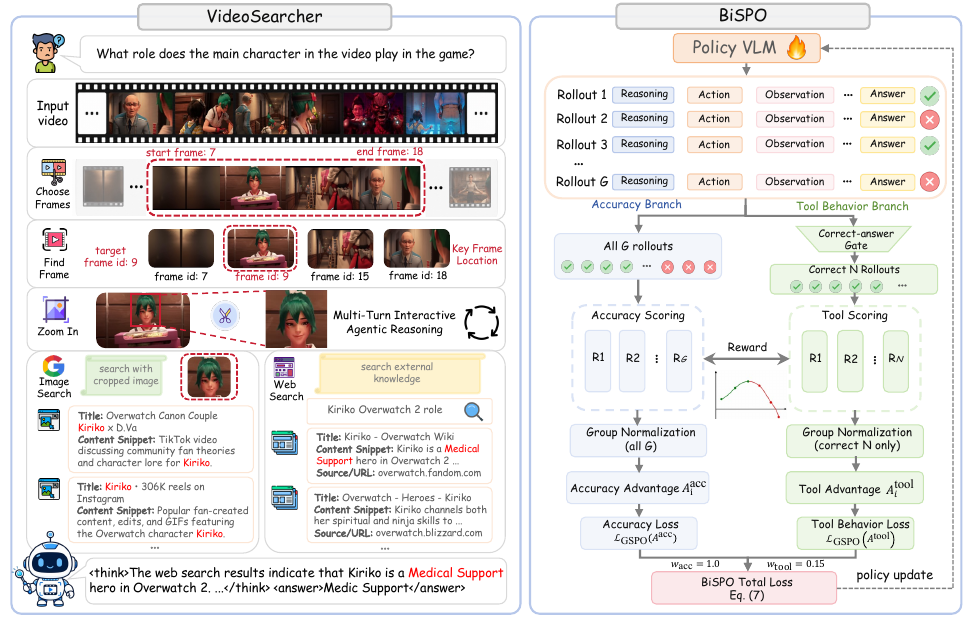} 
  \caption{Overview of the VideoSearcher framework with BiSPO training. VideoSearcher performs multi-turn reasoning by selecting relevant frames, localizing key evidence, zooming into target regions, and invoking multimodal search tools in VDR. BiSPO optimizes the policy VLM with complementary accuracy and tool-behavior branches.}
  \label{fig:framework}
\end{figure}
\subsection{Model Training}
We train VideoSearcher in two stages: cold-start SFT initializes the VDR interaction protocol and basic tool-use behaviors, while online RL further improves the long-horizon evidence acquisition, tool coordination, and reasoning capabilities. The framework of VideoSearcher and the RL algorithm is shown in Fig. \ref{fig:framework}.

\subsubsection{Cold-start SFT}
Given the curated trajectory set \(\mathcal{D}_{\mathrm{SFT}}=\{(x_i,\tau_i^\star)\}_{i=1}^{M}\), where \(x_i=(\mathcal{V}_i,q_i)\) denotes the question and \(\tau_i^\star\) is a validated multi-turn trajectory, we perform supervised fine-tuning with \(\mathcal{L}_{\mathrm{SFT}}=-\sum_{i=1}^{M}\log \pi_\theta(\tau_i^\star\mid x_i)\). Each trajectory contains interleaved reasoning steps, tool calls, tool observations (excluded from the loss), and the final answer. This stage initializes the VDR interaction protocol and basic tool-use behaviors before RL.

\subsubsection{RL with Bi-branch Sequence Policy Optimization} 
After cold-start SFT, we further optimize VideoSearcher with RL in the same VDR environment, and introduce Bi-branch Sequence Policy Optimization (BiSPO) on top of GSPO~\cite{zheng2025group} to decouple answer optimization from tool-behavior optimization. For each prompt \(x\), the old policy \(\pi_{\theta_{\mathrm{old}}}\) samples a group of \(G\) trajectories \(\{\tau_i\}_{i=1}^{G}\), where each trajectory contains interleaved reasoning steps, tool interactions, observations, and a final answer. Let \(y_i\) denote the model-generated token sequence in \(\tau_i\), with tool observations used only as conditioning context.

\noindent \textbf{Bi-branch Reward Design.} Successful VDR trajectories should produce correct answers while using tools purposefully. To preserve effective tool-use signals and promote purposeful tool invocation in VDR, we define two reward branches. The accuracy branch evaluates answer correctness and protocol validity:
\begin{equation}
R_i^{\mathrm{acc}} = R_i^{\mathrm{judge}} + \lambda_f R_i^{\mathrm{fmt}} + \lambda_d R_i^{\mathrm{dep}},
\end{equation}
where \(R_i^{\mathrm{judge}}\) is an LLM-judge score for final-answer correctness, \(R_i^{\mathrm{fmt}}\) checks format compliance, and \(R_i^{\mathrm{dep}}\) verifies correct tool-dependency constraints (e.g., selecting a frame before invoking region-level inspection or image search).




The tool branch is correctness-gated, so tool behavior is optimized only for trajectories that reach a correct answer:
\begin{equation}
\begin{adjustbox}{max width=\textwidth}
$\displaystyle
R_i^{\mathrm{tool}} = \mathbb{I}\!\left[R_i^{\mathrm{judge}} > \delta\right]\cdot
\operatorname{clip}\!\left(b+\gamma \sqrt{T_i^{\mathrm{vid}}}+\sum_{m \in \mathcal{M}}\sum_{k=1}^{U_{i,m}}\Delta_k,\ r_{\min},\ 1\right).
$
\end{adjustbox}
\end{equation}
where \(T_i^{\mathrm{vid}}\) counts valid video-grounding actions, \(\mathcal{M}=\{\mathrm{image},\mathrm{web}\}\) denotes search modalities, and \(U_{i,m}\) is the number of unique successful searches under modality \(m\). 
The marginal term \(\Delta_k\) induces a bell-shaped search utility per modality: early unique useful searches increase the tool reward, whereas excessive ones receive negative marginal rewards. Failed, empty, or duplicated calls receive zero marginal reward and do not advance the schedule.
This design encourages sufficient tool use in knowledge-intensive VDR while suppressing excessive or uninformative tool invocations.  See Appendix~\ref{appendix:b} for details.

\noindent \textbf{Decoupled Advantage Estimation.} For the accuracy branch, we compute group-relative advantages over all rollouts:
\begin{equation}
A_i^{\mathrm{acc}} = \frac{R_i^{\mathrm{acc}}-\mu^{\mathrm{acc}}}{\sigma^{\mathrm{acc}}+\epsilon},
\end{equation}
where $\mu^{\mathrm{acc}}$ and $\sigma^{\mathrm{acc}}$ are the mean and standard deviation of $\{R_j^{\mathrm{acc}}\}_{j=1}^{G}$. While  for the tool branch, comparison is restricted to correct trajectories. Let \(\mathcal{Q}=\{j \mid R_j^{\mathrm{judge}}>\delta\}\). The tool advantage is:
\begin{equation}
\begin{adjustbox}{max width=\textwidth}
$\displaystyle
A_i^{\mathrm{tool}} =
\mathbf{1}\{i\in\mathcal{Q},\ |\mathcal{Q}|\ge 2\}
\frac{R_i^{\mathrm{tool}}-\mu_{\mathcal{Q}}^{\mathrm{tool}}}{\sigma_{\mathcal{Q}}^{\mathrm{tool}}+\epsilon}.
$
\end{adjustbox}
\end{equation}
where $\mu_{\mathcal{Q}}^{\mathrm{tool}}$ and $\sigma_{\mathcal{Q}}^{\mathrm{tool}}$ are computed over $\{R_k^{\mathrm{tool}}\}_{k \in \mathcal{Q}}$. This avoids comparing tool behaviors from incorrect rollouts with evidence-seeking behaviors that actually support correct answers. When fewer than two rollouts are correct, we skip tool-behavior optimization and update only the accuracy branch. Once multiple correct trajectories emerge, the tool branch compares them to favor more effective tool use.

\begin{table*}[ht]
\centering
\footnotesize
\renewcommand{\arraystretch}{1.18}
\setlength{\tabcolsep}{4.5pt}
\resizebox{\textwidth}{!}{
\begin{tabular}{l|cccccccc|c}
\hline\hline
\textbf{Model} & \textbf{VideoDR} & \textbf{VSQA} & \textbf{MMSearch} & \textbf{HR-MMSearch} & \textbf{FVQA-test} & \textbf{InfoSeek} & \textbf{SimpleVQA} & \textbf{LiveVQA} & \textbf{Avg.} \\
\hline

\multicolumn{10}{c}{\textit{\textbf{Direct Answer}}} \\
\hline
\rowcolor{softorange}
\multicolumn{1}{l|}{\textit{Open-source}} & \multicolumn{8}{c|}{} & \\
Qwen3-VL-4B-Instruct & 8.00 & 12.59 & 13.45 & 3.61 & 25.33 & 25.75 & 46.89 & 23.35 & 19.87 \\
Qwen2.5-VL-7B-Instruct & 5.00 & 9.35 & 7.60 & 0.58 & 26.28 & 31.95 & 47.88 & 19.63 & 18.53 \\
Qwen3-VL-8B-Instruct & 8.00 & 16.91 & 11.70 & 12.13 & 24.22 & 23.15 & 42.94 & 23.18 & 20.28 \\
Qwen2.5-VL-32B-Instruct & 16.00 & 13.67 & 11.70 & 3.93 & 30.50 & 36.65 & 48.57 & 21.40 & 22.80 \\
Qwen3-VL-32B-Instruct & 16.00 & 20.86 & 16.96 & 19.02 & 32.17 & 28.95 & 45.90 & 31.59 & 26.43 \\

\hline
\rowcolor{softred}
\multicolumn{1}{l|}{\textit{Proprietary}} & \multicolumn{8}{c|}{} & \\
GPT-4o & 35.00 & 36.33 & 23.39 & 13.11 & 48.00 & 52.90 & 51.73 & 28.18 & 36.08 \\
Gemini-3-Flash & 57.00 & 56.12 & 57.31 & 21.97 & 56.50 & 54.85 & 63.57 & 38.90 & 50.78 \\
GPT-5.2 & 40.00 & 38.85 & 43.27 & 24.92 & 50.94 & 50.40 & 59.92 & 47.00 & 44.41 \\
Gemini-3-Pro & 61.00 & 54.32 & 62.57 & 26.89 & 59.22 & 56.30 & 64.07 & 40.06 & 53.05 \\

\hline
\multicolumn{10}{c}{\textit{\textbf{Agentic Model (zero-shot)}}} \\
\hline
\rowcolor{softorange}
\multicolumn{1}{l|}{\textit{Open-source}} & \multicolumn{8}{c|}{} & \\
Qwen3-VL-4B-Instruct & 25.00 & 29.14 & 49.12 & 24.92 & 31.72 & 28.10 & 41.95 & 26.15 & 32.01 \\
Qwen2.5-VL-7B-Instruct & 7.00 & 11.15 & 32.16 & 19.34 & 36.00 & 28.80 & 42.35 & 22.52 & 24.92 \\
Qwen3-VL-8B-Instruct & 28.00 & 30.94 & 47.37 & 27.87 & 53.61 & 46.15 & 62.29 & 39.37 & 41.95 \\
Qwen2.5-VL-32B-Instruct & 30.00 & 29.14 & 49.71 & 33.44 & 52.22 & 50.10 & 65.15 & 42.17 & 43.99 \\
Qwen3-VL-32B-Instruct & 36.00 & 37.77 & 49.12 & 34.43 & 54.28 & 49.85 & 64.17 & 42.87 & 46.06 \\

\hline
\rowcolor{softred}
\multicolumn{1}{l|}{\textit{Proprietary}} & \multicolumn{8}{c|}{} & \\
GPT-4o & 48.00 & 49.64 & 49.12 & 30.16 & 66.34 & 59.55 & 63.67 & 40.09 & 50.82 \\
Gemini-3-Flash & 60.00 & 58.99 & 62.57 & 41.64 & 64.89 & 61.10 & 67.92 & 48.06 & 58.15 \\
GPT-5.2 & 58.00 &  57.19  & 66.08 & {48.20} & {68.78} & {65.55} & {78.18} & {65.99} & {63.50} \\
Gemini-3-Pro & 77.00 & 65.11 & {74.27} & {48.52} & {72.61} & {66.45} & {75.91} & {59.69} & {67.45} \\

\hline
\multicolumn{10}{c}{\textit{\textbf{Agentic Model}}} \\
\hline
Visual-ARFT & 5.00 & 8.99 & 34.50 & 24.92 & 41.72 & 37.95 & 42.45 & 25.40 & 27.62 \\
Webwatcher-7B & 37.00 & 34.53 & 49.10 & 26.89 & 58.17 & \underline{56.90} & 54.30 & \underline{51.20} & 46.01 \\
MMSearch-R1 & 5.00 & 11.87 & 53.80 & 20.33 & 58.40 & 55.10 & 57.40 & 48.40 & 38.79 \\
DeepEyesV2 & 19.00 & 21.58 & 63.70 & 14.10 & \underline{60.60} & 51.10 & 59.40 & 24.43 & 39.24 \\

\rowcolor{softblue}
\textbf{VideoSearcher-4B} & \underline{46.00} & \underline{49.28} & \underline{66.08} & \underline{39.67} & 55.17 & 49.25 & \underline{65.84} & 50.72 & \underline{52.75} \\

\color{softgreen}$\Delta$ v.s. Qwen3-VL-4B-Instruct
& \color{softgreen}\textbf{+21.00} 
& \color{softgreen}\textbf{+20.14} 
& \color{softgreen}\textbf{+16.96} 
& \color{softgreen}\textbf{+14.75} 
& \color{softgreen}\textbf{+23.45} 
& \color{softgreen}\textbf{+21.15} 
& \color{softgreen}\textbf{+23.89} 
& \color{softgreen}\textbf{+24.57} 
& \color{softgreen}\textbf{+20.74} \\

\rowcolor{softblue}
\textbf{VideoSearcher-8B} & \textbf{53.00} & \textbf{51.80} & \textbf{67.84} & \textbf{43.61} & \textbf{64.10} & \textbf{58.30} & \textbf{67.20} & \textbf{55.40} & \textbf{57.66} \\

\color{softgreen}$\Delta$ v.s. Qwen3-VL-8B-Instruct
& \color{softgreen}\textbf{+25.00} 
& \color{softgreen}\textbf{+20.86} 
& \color{softgreen}\textbf{+20.47} 
& \color{softgreen}\textbf{+15.74} 
& \color{softgreen}\textbf{+10.49} 
& \color{softgreen}\textbf{+12.15} 
& \color{softgreen}\textbf{+4.91} 
& \color{softgreen}\textbf{+16.03} 
& \color{softgreen}\textbf{+15.71} \\

\hline\hline
\end{tabular}
}
\caption{Performance on search-oriented benchmarks under Direct Answer and Agentic Model workflows. Zero-shot denotes tool-augmented inference without task-specific training. Scores are averaged over benchmark samples.}
\label{tab:agentic_search}
\end{table*}

\noindent \textbf{BiSPO Objective.} Long VDR trajectories contain many reasoning tokens and tool-conditioned contexts, making token-level clipping unstable. We therefore build BiSPO on sequence-level importance sampling. For trajectory \(i\), the sequence importance ratio is:
\begin{equation}
\begin{adjustbox}{max width=\textwidth}
$\displaystyle
s_i(\theta)=\exp\!\left(\bar{\ell}_i(\theta)\right),\quad
\bar{\ell}_i(\theta)=\frac{1}{|y_i|}\sum_t\!\left[
\log \pi_\theta\!\left(y_{i,t}\mid x_i,y_{i,<t}\right)-
\log \pi_{\theta_{\mathrm{old}}}\!\left(y_{i,t}\mid x_i,y_{i,<t}\right)
\right].
$
\end{adjustbox}
\end{equation}
Given an advantage $A$, the clipped sequence-level surrogate objective is:
\begin{equation}
\begin{adjustbox}{max width=\textwidth}
$\displaystyle
\mathcal{L}_{\mathrm{GSPO}}(A)=-\mathbb{E}_i\!\left[
\min\!\left(s_i(\theta)A_i,\,\tilde{s}_i(\theta)A_i\right)
\right],\quad
\tilde{s}_i(\theta)=\operatorname{clip}\!\left(s_i(\theta),1-\epsilon_{\mathrm{low}},1+\epsilon_{\mathrm{high}}\right).
$
\end{adjustbox}
\end{equation}
The final BiSPO objective combines the two branches only at the loss level:
\begin{equation}
\mathcal{L}_{\mathrm{BiSPO}} =
w_{\mathrm{acc}}\mathcal{L}_{\mathrm{GSPO}}\!\left(A^{\mathrm{acc}}\right)
+w_{\mathrm{tool}}\mathcal{L}_{\mathrm{GSPO}}\!\left(A^{\mathrm{tool}}\right).
\end{equation}
We set $w_{\mathrm{acc}}=1.0$ and $w_{\mathrm{tool}}=0.15$. During RL, the vision tower is frozen and the actor is optimized with sequence-level aggregation. By separating answer correctness from tool behavior, BiSPO preserves the primary task objective while providing a dedicated learning signal for efficient temporal grounding, multimodal retrieval, and evidence-aware stopping for VDR.


\section{Experiment}


\subsection{Implementation Details}
\noindent \textbf{Model and Training.} VideoSearcher is initialized from Qwen3-VL-8B-Instruct and trained with a two-stage pipeline, using LLaMA-Factory \cite{zheng2024llamafactory} for SFT and veRL \cite{sheng2025hybridflow} for online RL. In SFT, we freeze the vision encoder and multi-modal projector and fine-tune only the language model with a learning rate of $1\times10^{-5}$. The RL stage starts from the SFT checkpoint and is trained for one epoch, with the vision tower kept frozen. We use a prompt batch size of 64, sample 4 rollouts per prompt, and set the actor learning rate to $1\times10^{-5}$. More implementation details about hyperparameter setting are provided in the Appendix \ref{appendix:b}.


\noindent \textbf{Benchmarks.} To comprehensively evaluate our proposed VideoSearcher, we consider three categories of benchmarks. For Video Deep Research, we use VideoDR~\cite{liu2026watching} and our newly constructed VideoSearch-QA (VSQA). For multimodal search, we evaluate on MMSearch~\cite{jiang2024mmsearch}, HR-MMSearch~\cite{chng2025sensenova}, FVQA-test~\cite{wu2025mmsearch}, InfoSeek~\cite{chen2023can}, SimpleVQA~\cite{cheng2025simplevqa}, and LiveVQA~\cite{fu2025livevqa}. For general video understanding, we further test on MMVU~\cite{zhao2025mmvu}, TempCompass~\cite{liu2024tempcompass}, VideoMMMU~\cite{hu2025video}, and VideoMathQA~\cite{rasheed2025videomathqa}. Additional benchmark details are provided in the Appendix \ref{appendix:eval_bench}.

\noindent \textbf{Baselines.} We compare VideoSearcher with a broad set of advanced baseline models. Proprietary baselines include GPT-4o~\cite{hurst2024gpt}, GPT-5.2~\cite{singh2025openai}, and Gemini-3-Flash/Pro~\cite{pichai2025new, gemini3pro}, while the powerful open-source VLMs include the Qwen2.5-VL~\cite{bai2025qwen2} and Qwen3-VL~\cite{bai2025qwen3} series. We further compare with multimodal search-oriented agentic models, including Visual-ARFT~\cite{liu2025visual}, Webwatcher~\cite{geng2025webwatcher}, MMSearch-R1~\cite{wu2025mmsearch}, and DeepEyesV2~\cite{hong2025deepeyesv2}, as well as the video reasoning models, including Video-R1~\cite{feng2026video}, VideoChat-R1.5~\cite{yan2026videochat}, VideoRFT~\cite{wang2026videorft}, and VideoCoM~\cite{rasheed2025video}. For search-oriented benchmarks, models are evaluated under two settings: (1) \textit{Direct Answer}, where the model answers directly without using external tools, and (2) \textit{Agentic Model}, where the model is provided with the available tools and autonomously decides how to invoke them during rollout reasoning. For video understanding tasks, baselines use uniformly sampled frames in the Direct Answer setting, while VideoSearcher is evaluated as an Agentic Model.

\subsection{Main Results}
\noindent \textbf{Search-oriented Benchmarks.} As shown in Tab.~\ref{tab:agentic_search}, our method achieves strong performance across search-oriented evaluations. On VideoDR and VSQA, VideoSearcher substantially improves over existing agentic baselines, demonstrating its effectiveness for VDR scenarios that require joint video grounding, tool use, and open-web evidence reasoning. Notably, although trained only on video-based data, VideoSearcher also generalizes well to general multimodal search benchmarks, substantially outperforming Qwen3-VL agentic baselines.


\begin{table}[ht]
\centering
\footnotesize
\renewcommand{\arraystretch}{1.15}
\setlength{\tabcolsep}{5pt}
\resizebox{0.94\linewidth}{!}{
\begin{tabular}{l|cccc|c}
\hline\hline
\textbf{Model} & \textbf{MMVU} & \textbf{TempCompass} & \textbf{VideoMMMU} & \textbf{VideoMathQA} & \textbf{Avg.} \\
\hline
Qwen3-VL-4B-Instruct & {63.36} & {70.95} & {60.78} & 22.62 & {54.43} \\
Qwen3-VL-8B-Instruct & \underline{69.28} & \underline{73.62} & \underline{65.67} & 26.90 & {58.87} \\
Video-R1 & 63.80 & 73.20 & 52.40 & 23.30 & 53.18 \\
VideoChat-R1.5 & 62.08 & 72.30 & 51.40 & 25.70 & 52.87  \\
VideoRFT & {68.50} & \textbf{73.70} & 51.10 & 25.20 & 54.63 \\
VideoCoM & 65.40 & 71.30 & 50.20 & {27.80} & 53.68 \\
\hline
\rowcolor{softblue}
\textbf{VideoSearcher-4B} & 68.64 & 70.27 & 65.33 & \underline{27.86} & \underline{58.03} \\
\rowcolor{softblue}
\textbf{VideoSearcher-8B} & \textbf{70.40} & 73.14 & \textbf{66.33} & \textbf{31.67} & \textbf{60.39} \\
\hline\hline
\end{tabular}
}
\caption{Performance on general video understanding benchmarks.}
\label{tab:video_understanding}
\end{table}

\noindent \textbf{General Video Understanding.} Tab.~\ref{tab:video_understanding} further evaluates VideoSearcher on general video understanding benchmarks. Despite being designed as a search-enabled agent, where retrieved evidence may introduce distracting or irrelevant information, VideoSearcher  maintains competitive general video reasoning ability. The experimental results indicate that the proposed agentic training does not merely optimize search behavior, but also preserves and improves transferable video understanding under more complex evidence conditions.

\subsection{Ablation Study}

\begin{table}[ht]
\centering
\scriptsize
\renewcommand{\arraystretch}{1.12}
\setlength{\tabcolsep}{3.5pt}
\resizebox{0.82\linewidth}{!}{
\begin{tabular}{l|l|cccc}
\hline\hline
\textbf{Method} & \textbf{Setting} & \textbf{VideoDR} & \textbf{VSQA} & \textbf{MMSearch} & \textbf{HR-MMSearch} \\
\hline
SFT  & SFT only & 46.00 & 47.84 & 58.48 & 40.98 \\
GRPO & $R_{\mathrm{acc}}$ only & 45.00 & 48.16 & 65.50 & 41.14 \\
GSPO & $R_{\mathrm{acc}}$ only & 47.00 & 50.72 & 63.16 & 40.66 \\
GSPO & $R_{\mathrm{acc}}+0.15R_{\mathrm{tool-mono}}$ & 47.00 & 51.08 & 65.50 & 40.33 \\
BiSPO & $R_{\mathrm{acc}}+0.15R_{\mathrm{tool-mono}}$ & 50.00 & 51.44 & 65.50 & 40.33 \\
\rowcolor{softblue}
\textbf{BiSPO} & \textbf{Full} & \textbf{53.00} & \textbf{51.80} & \textbf{67.84} & \textbf{43.61} \\
\hline\hline
\end{tabular}
}
\caption{Algorithmic ablation of RL. GRPO and GSPO are single-branch RL variants; and BiSPO is the proposed bi-branch variant, with BiSPO Full using the correctness-gated bell-shaped tool reward.}
\label{tab:algorithm_ablation}
\end{table}

\noindent \textbf{RL Ablation Study.} Tab.~\ref{tab:algorithm_ablation} ablates the RL optimization design. With only the accuracy reward $R_{\mathrm{acc}}$, GSPO improves over GRPO on VideoDR and VSQA, reflecting the benefit of sequence-level optimization for long VDR trajectories. However, its mixed VDR gains suggest that answer-level supervision alone does not reliably induce tool use. We further compare a monotonic tool reward, $R_{\mathrm{tool\text{-}mono}}$, which rewards every valid tool call. Our full reward instead follows a bell-shaped design, encouraging sufficient grounding and retrieval while penalizing unnecessary calls. By decoupling $R_{\mathrm{acc}}$ from the tool reward, BiSPO  optimizes tool use in reasoning-intensive VDR tasks, improving VideoDR from 47.00\% to 50.00\% with $R_{\mathrm{tool\text{-}mono}}$ and achieving the best overall results with the full bell reward. These results show that decoupled optimization and bell-shaped tool reward are both important for stable and effective VDR training.


\begin{table}[ht]
\centering
\footnotesize
\renewcommand{\arraystretch}{1.15}
\setlength{\tabcolsep}{6pt}
\resizebox{0.82\linewidth}{!}{
\begin{tabular}{cc|cccc}
\hline\hline
\textbf{Locate} & \textbf{Search} & \textbf{VideoDR} & \textbf{VSQA} & \textbf{MMSearch} & \textbf{HR-MMSearch} \\
\hline
$\checkmark$ &  & 12.00 & 19.78 & 12.28 & 3.93 \\
 & $\checkmark$ & 43.00 & 47.48 & 66.67 & 40.98 \\
\rowcolor{softblue}
$\checkmark$ & $\checkmark$ & \textbf{53.00} & \textbf{51.80} & \textbf{67.84} & \textbf{43.61} \\
\hline\hline
\end{tabular}
}
\caption{Tool ablation study. Locate denotes the video localization tools and Search denotes the external web retrieval tools.}
\label{tab:tool_ablation}
\end{table}

\noindent \textbf{Tool Ablation Study.} Tab. \ref{tab:tool_ablation} ablates the localization and search tools. Localization alone performs poorly, indicating that grounded visual evidence is insufficient for open-world questions without external knowledge. Search alone also degrades markedly, especially when retrieval requires prior grounding in video-centric benchmarks. These results show that the two tools are complementary: localization anchors the query in visual evidence, while search augments it with external information for knowledge-intensive reasoning.



\section{Conclusion}
In this work, we study Video Deep Research (VDR), which extends video understanding from closed-context perception to open-world information exploration. We introduce {VideoSearcher}, a closed-loop agentic framework for temporal grounding, spatial inspection, and multimodal retrieval over dynamic videos. We further develop a video-centric data pipeline for synthesizing high-quality multi-tool trajectories and propose {BiSPO}, a dual-branch RL algorithm that decouples answer accuracy from tool-invocation optimization for knowledge-intensive VDR. Finally, we introduce {VideoSearch-QA}, a benchmark for evaluating video information grounding and multimodal search-based reasoning in open-world scenarios. Extensive experiments across multiple benchmarks demonstrate the effectiveness of VideoSearcher. 
\section*{Limitations}
Although VideoSearcher improves search-oriented video question answering, several limitations remain. It relies on an external tool environment, so answer quality depends on the latency, coverage, and stability of image/web search, making reproduction less deterministic than closed-book video QA. Online RL with multi-turn tool use is also computationally expensive, as long contexts and repeated rollouts increase both training cost and engineering complexity. In addition, our training and evaluation mainly focus on search-oriented visual/video QA, and may not fully cover tasks requiring dense temporal localization, audio understanding, embodied interaction, specialized domain knowledge, or short-event motion cues under the 1-fps frame representation. Finally, open-ended evaluation relies on an LLM judge, which can introduce noise for ambiguous or semantically equivalent answers; stronger human evaluation and calibrated automatic metrics remain useful future directions.




\bibliography{custom}

\clearpage

\appendix
{\LARGE\textbf{Appendix}\par}\normalsize
\setlength{\floatsep}{8pt plus 2pt minus 2pt}
\setlength{\textfloatsep}{10pt plus 2pt minus 2pt}
\setlength{\intextsep}{8pt plus 2pt minus 2pt}
\setlength{\parskip}{2pt plus 1pt minus 1pt}
\makeatletter
\setlength{\@fptop}{0pt}
\setlength{\@fpsep}{10pt plus 2pt minus 2pt}
\setlength{\@fpbot}{0pt plus 1fil}
\makeatother
\captionsetup{skip=4pt}
\tcbset{
  appendixprompt/.style={
    colback=gray!5,
    colframe=gray!60!black,
    boxrule=0.6pt,
    fonttitle=\bfseries,
    left=0.6em,
    right=0.6em,
    top=0.55em,
    bottom=0.55em,
    width=0.96\textwidth,
    arc=4pt
  }
}
\small
\section{Data Pipeline}
\subsection{Training Data Construction}
\label{appendix:a.1}
We provide more details on the construction of the training data used for cold-start SFT and online RL. As illustrated in Fig.~\ref{fig:data_pipeline}, our goal is to convert knowledge-intensive image--text QA samples into video-centric deep-research instances, where the model must ground visual clues in a video, invoke appropriate tools, retrieve external evidence, and synthesize a final answer.

\noindent \textbf{Seed QA Collection.} We begin with knowledge-intensive image--text QA data from existing multimodal search and visual information-seeking datasets. These samples are suitable seeds because their answers usually cannot be obtained from visual perception alone, but require external knowledge from web pages or image search. In addition to existing datasets, we also include human-annotated seed questions to increase coverage over diverse entities and search scenarios. For each seed sample, we retain the question, answer, reference image, and the key visual entity that supports the answer.

\noindent \textbf{Entity-driven Video Acquisition.} Since the original seed samples are static image--text pairs, we first identify the core visual entity that should appear in a video, such as a landmark, logo, product, person, scene, or event-related object. To reduce entity ambiguity, we query two strong multimodal models, Qwen3.5-397B~\cite{qwen3.5} and Kimi-K2.5~\cite{team2026kimi}, and only keep entities that are consistently recognized by both models. The verified entities are then used as retrieval targets to collect relevant videos from multiple platforms, including web video sources such as YouTube and Bilibili. This produces an initial pool of video-QA candidates grounded in real video content.

\noindent \textbf{Video Normalization and QA-alignment.} Collected videos are normalized into a unified format before trajectory synthesis. We resample each video into frame sequences and add explicit frame indices as temporal references, enabling subsequent tools to localize and refer to specific moments. We then rewrite the original image-based questions into video-grounded questions. During rewriting, we ensure that the question naturally refers to the video context rather than the original image, and we add discriminative visual or temporal descriptions when multiple similar entities appear in the same video. We further apply visual-QA alignment verification to remove samples where the target entity is missing, the answer is directly exposed by OCR, the video contains inconsistent visual anchors, or the question can be answered without using the video evidence.

\noindent \textbf{Hard Sample Mining.} To avoid constructing a training set dominated by visually obvious or shallow-search examples, we perform hard sample mining with Qwen3-VL-8B-Instruct. For each candidate video-QA pair, the model is rolled out multiple times under the agentic setting. We retain samples that the model fails to solve consistently, following the criterion that the number of successful rollouts is no more than one. This step encourages the final data to focus on challenging cases that require temporal localization, fine-grained visual inspection, multimodal retrieval, and evidence-grounded reasoning. The resulting samples are then split into two branches: 3,285 video instances are reserved for online RL rollouts, while the remaining samples are used for SFT trajectory synthesis.

\noindent \textbf{Hierarchical Trajectory Synthesis.} For each SFT sample, we synthesize a complete multi-tool reasoning trajectory with a hierarchical teacher pipeline. A lightweight closed-source model is first used for coarse temporal screening through \texttt{choose\_frames}, which narrows the video to a smaller set of potentially relevant frames. A more capable closed-source model then performs fine-grained localization and reasoning, including \texttt{find\_frame}, optional \texttt{zoom\_in}, \texttt{image\_search}, and \texttt{web\_search}. The generated trajectory records the complete interaction process, including intermediate reasoning states, tool calls, tool observations, retrieved evidence, and the final answer. To reduce visual token cost, we follow an incremental visual-context strategy: newly acquired visual observations are introduced at each turn, while the full textual reasoning history is preserved.

\noindent \textbf{Quality Control.} We apply strict filtering before adding synthesized trajectories to the cold-start dataset. Rule-based filters remove trajectories with malformed tool calls, invalid frame references, illegal bounding boxes, failed searches, repeated queries, or repetitive reasoning patterns. We further use Kimi-K2.5 as a quality verifier to reject trajectories with hallucinated evidence, unsupported conclusions, inconsistent tool observations, or broken reasoning chains. After filtering, we obtain 3,811 high-quality SFT trajectories. These trajectories provide diverse demonstrations of video localization, spatial inspection, multimodal search, and evidence synthesis, thereby initializing the interaction protocol and basic tool-use behaviors of VideoSearcher before RL.
\begin{figure*}[htbp]
    \centering
    \includegraphics[width=0.85\linewidth]{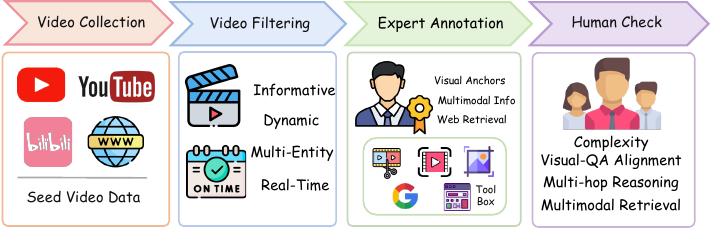}
    \caption{VideoSearch-QA Benchmark Construction Pipeline}
    \label{fig:vsqa_pipeline}
\end{figure*}
\subsection{VideoSearch-QA Benchmark Construction}
\label{appendix:a.2}
We construct VideoSearch-QA (VSQA) as a human-centered benchmark for evaluating video-grounded multimodal search. Unlike existing VDR evaluation that often converts visual cues into text before retrieval, VSQA is designed to test whether an agent can directly identify visual anchors from dynamic videos, use these anchors to perform web or image search, and reason over multimodal evidence. As shown in Tab. \ref{tab:vsqa-domain-distribution}, the final benchmark contains 278 manually annotated samples across diverse domains, including landmarks, geography, natural scenes, culture, sports and gaming, leisure, technology, and business. Fig. \ref{fig:vsqa_pipeline} illustrates the VSQA construction pipeline.

\noindent \textbf{Video Collection and Filtering.} We collect candidate videos from public web video platforms and online sources, prioritizing videos that are informative, dynamic, multi-entity, and temporally grounded. In particular, we favor videos containing recognizable and searchable visual anchors, such as landmarks, logos, signs, products, public figures, maps, or event-specific scenes. We also include recent or real-time videos, especially from 2026 events, to reduce potential memorization from model pretraining. Human annotators then screen the collected videos and remove cases that are visually ambiguous, low-resolution, dominated by irrelevant content, or unsuitable for open-web retrieval. The retained videos must contain at least one stable visual anchor, require localized video grounding, and cannot be answered solely from the video or solely from web evidence without using video cues.

\noindent \textbf{Human Annotation and Verification.}  For each retained video, expert annotators manually construct the question-answer pair based on key visual cues in the video. The question is designed to be grounded in concrete video evidence, such as a visible object, text region, location, person, logo, or event clue, while requiring multimodal retrieval through web search or image search to obtain the final answer. Annotators also provide the reference answer and verify it with reliable external evidence. A separate human checking stage further reviews each video-question-answer triple in terms of visual-QA alignment, difficulty, multi-hop reasoning, and multimodal retrieval necessity. Samples with ambiguous visual anchors, recognition-only answers, unsupported answers, or insufficient need for external retrieval are revised or discarded.

\begin{table}[htbp]
\centering
\small
\setlength{\tabcolsep}{4pt}
\begin{adjustbox}{width=0.8\linewidth}
\begin{tabularx}{\columnwidth}{@{}>{\raggedright\arraybackslash}Xcc@{}}
\toprule
\textbf{Domain} & \textbf{\# Samples} & \textbf{Percentage} \\
\midrule
Landmarks & 65 & 23.4\% \\
Geography & 51 & 18.3\% \\
Natural Scenes & 31 & 11.2\% \\
Culture & 47 & 16.9\% \\
Sports \& Games & 27 & 9.7\% \\
Leisure & 25 & 9.0\% \\
Technology & 18 & 6.5\% \\
Business & 14 & 5.0\% \\
\midrule
\textbf{Summary} & \textbf{278} & \textbf{100.0\%} \\
\bottomrule
\end{tabularx}
\end{adjustbox}
\caption{Data distribution of the VideoSearch-QA benchmark.}
\label{tab:vsqa-domain-distribution}
\end{table}

\FloatBarrier

\section{Implementation Details}
\label{appendix:b}
\subsection{Training Setting}
VideoSearcher is trained with a two-stage pipeline. The first stage is cold-start supervised fine-tuning on curated multi-turn Video Deep Research trajectories. These trajectories follow the same interaction format used during inference, including intermediate reasoning, tool invocation, tool observation, and final answer generation. During SFT, we freeze the vision encoder and multi-modal projector and update only the language model.

The second stage is online reinforcement learning in the tool-interactive environment. The RL stage is initialized from the SFT checkpoint and trained for one epoch. The model can interact with five tools: \texttt{choose\_frames}, \texttt{find\_frame}, \texttt{zoom\_in}, \texttt{image\_search}, and \texttt{web\_search}. During RL, the vision tower remains frozen to stabilize long-horizon multi-turn training.

\subsection{Hyperparameter Setting}
For SFT, we use bf16 precision and a learning rate of $1\times10^{-5}$. The vision encoder and multi-modal projector are frozen, while the language model is optimized.

For RL, we use a prompt batch size of 64 and sample 4 rollouts for each prompt. The actor learning rate is $1\times10^{-5}$. The maximum response length is 16{,}384 tokens, and the maximum context length is 49{,}152 tokens. The maximum number of assistant turns is 12. We use asymmetric clipping with $\epsilon_{\mathrm{low}}=0.2$ and $\epsilon_{\mathrm{high}}=0.28$. The answer-correctness branch has weight 1.0, and the tool-behavior branch has weight 0.15. We do not use an additional KL reward or KL loss during RL.

For the reward design, we instantiate the answer-correctness branch by absorbing the coefficients in Eq.~(1) into the corresponding terms:
\[
R_i^{\mathrm{acc}} = R_i^{\mathrm{judge}} + R_i^{\mathrm{fmt}} + R_i^{\mathrm{dep}}.
\]
where \(R_i^{\mathrm{judge}}\in\{0,1\}\) is given by the LLM judge, \(R_i^{\mathrm{fmt}}=0.5\) if the trajectory satisfies the required output format and has no model-caused tool error, and 0 otherwise. The dependency term penalizes invalid tool ordering as \(R_i^{\mathrm{dep}}=-\min(0.1N_i^{\mathrm{dep}},0.5)\), where \(N_i^{\mathrm{dep}}\) is the number of detail/search tool calls made without a valid locked frame. The correctness gate threshold for the tool branch is \(\delta=0\).

For the tool-behavior branch, non-search video-grounding actions use \(b=0.3\) and \(\gamma=0.35\) in the \(b+\gamma\sqrt{T_i^{\mathrm{vid}}}\) term. For search-tool shaping, image search and web search are counted separately. For each modality, the marginal reward for the \(k\)-th unique useful search call is \([0.35,0.20,0.10,-0.10,-0.25]\) for \(k=1,\ldots,5\), and \(-0.60\) for every subsequent unique useful call. Failed calls, empty image-search results, duplicated text queries, and repeated visual regions receive zero marginal reward and do not advance the count. The final tool score is clipped to \([-1.0,1.0]\).

\subsection{Evaluation Setting}
Evaluation is conducted in a tool-interactive environment consistent with the capabilities required by each benchmark. For VideoDR and VSQA, which require video grounding and open-world evidence retrieval, the model is allowed to use all five tools: \texttt{choose\_frames}, \texttt{find\_frame}, \texttt{zoom\_in}, \texttt{image\_search}, and \texttt{web\_search}. For multimodal search benchmarks, where the input is a static image rather than a video, we provide the image-level tools \texttt{zoom\_in}, \texttt{image\_search}, and \texttt{web\_search}. For general video understanding benchmarks, we keep the same five-tool environment as VDR to evaluate whether the trained agent can generalize beyond search-oriented tasks. During evaluation, we use temperature 0.7, top-\(p=0.8\), and top-\(k=20\) for decoding, with a maximum of 12 assistant turns. We report average scores over all samples in each benchmark.

For open-ended benchmarks, we use Qwen3.5-27B as the automatic judge. This choice balances judgment quality and evaluation throughput: the model provides strong instruction-following ability while remaining efficient to serve as a dense 27B judge, which is important because each open-ended evaluation requires many independent judge calls. For multiple-choice benchmarks, we use exact-match accuracy.

\section{Baseline Models}
\subsection{Open-source Models}
We compare our method against a broad set of open-source VLMs and multimodal agentic models:
\begin{itemize}[leftmargin=*, itemsep=2pt, topsep=2pt, parsep=0pt]
    \item \textbf{Qwen-VL Series.} We evaluate Qwen2.5-VL~\cite{bai2025qwen2} and Qwen3-VL~\cite{bai2025qwen3} models as strong open-source generalist VLMs. Qwen2.5-VL supports dynamic-resolution visual encoding and temporal localization for image/video understanding, while Qwen3-VL further improves long-context multimodal reasoning, interleaved image-video-text understanding, and spatial-temporal modeling.
    \item \textbf{Visual-ARFT.} Visual-ARFT~\cite{liu2025visual} improves VLMs with agentic reinforcement fine-tuning. It equips models with external tool-use abilities, such as web browsing and image manipulation, making it a relevant baseline for visual agentic reasoning.
    \item \textbf{WebWatcher.} WebWatcher~\cite{geng2025webwatcher} is a multimodal deep-research agent trained with synthetic trajectories and reinforcement learning. It focuses on visual-textual information seeking with dynamic tool use, serving as a strong open-web research baseline.
    \item \textbf{MMSearch-R1.} MMSearch-R1~\cite{wu2025mmsearch} is an RL-based multimodal search agent that learns when and how to invoke text and image search tools. It is designed for on-demand, multi-turn search in real-world Internet environments.
    \item \textbf{DeepEyesV2.} DeepEyesV2~\cite{hong2025deepeyesv2} studies how to build agentic multimodal models with external tools such as web search and code execution. It emphasizes interleaved multimodal reasoning and tool-integrated problem solving.
    \item \textbf{Video-R1.} Video-R1~\cite{feng2026video} adapts the R1-style reinforcement learning paradigm to video reasoning. It introduces temporal-aware RL training to improve spatial, temporal, and logical reasoning over videos.
    \item \textbf{VideoChat-R1.5.} VideoChat-R1.5~\cite{yan2026videochat} introduces visual test-time scaling with iterative perception. It progressively refines spatio-temporal attention during inference to strengthen multimodal video reasoning.
    \item \textbf{VideoRFT.} VideoRFT~\cite{wang2026videorft} extends reinforced fine-tuning to video reasoning through SFT on video CoT data followed by RL. It further uses semantic-consistency rewards to align textual reasoning with visual evidence.
    \item \textbf{VideoCoM.} VideoCoM~\cite{rasheed2025video} proposes interactive video reasoning via a chain of tool-augmented manipulations. It treats video as an active reasoning workspace, where models iteratively manipulate and inspect visual evidence.
\end{itemize}
\subsection{Proprietary Models}
We also include several powerful proprietary models for evaluation:
\begin{itemize}[leftmargin=*, itemsep=2pt, topsep=2pt, parsep=0pt]
    \item \textbf{GPT-4o.} GPT-4o~\cite{hurst2024gpt} is a proprietary multimodal model with strong visual understanding, instruction following, and general reasoning ability.

    \item \textbf{GPT-5.2.} GPT-5.2~\cite{singh2025openai} is an advanced proprietary multimodal model with strong reasoning and tool-use capabilities. We evaluate it as a competitive closed-source baseline.

    \item \textbf{Gemini-3-Flash/Pro.} Gemini-3-Flash and Gemini-3-Pro~\cite{pichai2025new} are proprietary multimodal models from the Gemini family. Gemini-3-Flash emphasizes efficient inference, while Gemini-3-Pro provides stronger reasoning and multimodal understanding.
\end{itemize}

\section{Evaluation Benchmarks}
\label{appendix:eval_bench}
\subsection{Search-oriented  Benchmarks}
For evaluating Video Deep Research and multimodal image-text search capabilities, we primarily adopt the following benchmarks: 
\begin{itemize}[leftmargin=*, itemsep=2pt, topsep=2pt, parsep=0pt]
    \item \textbf{VideoDR.} VideoDR~\cite{liu2026watching} is an open-domain video deep research benchmark that evaluates whether agents can combine video-grounded clues with open-web evidence. The full benchmark contains 500 samples (released on 2026.05.19) across six domains: Daily Life, Technology, Culture, History, Economics, and Geography. Each question requires extracting multi-frame visual anchors, performing interactive web search, and synthesizing video--web evidence for a verifiable answer. In our evaluation, we use the 100-sample version released on 2026.01.14 from the official repository, which additionally provides Category and Difficulty labels. Overall, VideoDR evaluates the agentic ability of multimodal models to connect video-grounded clues with external web evidence for open-domain factual reasoning.
    \item \textbf{VideoSearch-QA (VSQA).} VideoSearch-QA is our manually annotated benchmark for evaluating video-grounded multimodal search. It contains 278 expert-annotated video--question pairs across eight domains, including Landmarks, Geography, Natural Scenes, Culture, Sports \& Games, Leisure, Technology, and Business. Each question is grounded in key visual cues from the video, such as objects, text regions, locations, logos, people, or event-specific clues, and requires external web or image search to obtain the final answer. VSQA evaluates whether agents can localize relevant visual evidence in dynamic videos, transform video-grounded clues into effective retrieval queries, and synthesize multimodal evidence for open-world factual reasoning.
    \item \textbf{MMSearch.} MMSearch~\cite{jiang2024mmsearch} consists of 300 manually curated examples spanning 14 subdomains, organized into two parts: News and Knowledge. The News split is designed around events after August 2024, reducing potential contamination from model pretraining, whereas the Knowledge split emphasizes obscure or long-tail facts that remain difficult even for strong models. Following MMSearch-R1, we restrict our evaluation to the 171 image-based questions and remove text-only samples, so that the benchmark better reflects visually grounded information-seeking scenarios.
    \item \textbf{HR-MMSearch.} HR-MMSearch~\cite{chng2025sensenova} contains 305 image–question pairs, built from 4K-resolution news images from 2025 to reduce potential overlap with pre-training data and provide rich visual details. The dataset spans eight diverse domains: Sports, Entertainment \& Culture, Science \& Technology, Business \& Finance, Games, Academic Research, Geography \& Travel, and Others. HR-MMSearch contains 188 Hard and 117 Easy samples. Each question is knowledge-intensive and grounded in key visual evidence, often involving small objects or text regions that occupy less than 5\% of the image. Solving these questions requires the use of at least one multimodal tool, including image search, text search, or image cropping. Overall, HR-MMSearch provides a challenging and diverse testbed for evaluating fine-grained visual understanding and agentic search in VLM agents.
    \item \textbf{FVQA-test.} FVQA-test~\cite{wu2025mmsearch} is a multimodal evaluation set designed to cover both visual and textual knowledge domains. It contains 1,800 high-quality examples from three sources. Specifically, 600 examples are drawn from FVQA-auto-vc after accuracy verification and training-data separation, 600 examples come from the InfoSeek Human Split with manually corrected answers, and the remaining 600 examples are newly annotated by humans to further broaden the benchmark coverage.
    \item \textbf{InfoSeek.} InfoSeek~\cite{chen2023can} is a real-world knowledge retrieval benchmark built from Wikidata triples. Its questions are generated by converting structured triples into natural-language queries with human-designed templates covering 300 relations, where unit and entity-type placeholders are incorporated to improve question clarity. The dataset is further refined by removing unanswerable questions and balancing samples across entities. In our setting, the evaluation subset contains 2,000 instances sampled from the test split, covering a diverse range of factual queries.
    \item \textbf{SimpleVQA.} SimpleVQA~\cite{cheng2025simplevqa} is a factual VQA benchmark designed to evaluate objective real-world knowledge. It consists of two types of samples: image–question pairs collected from post-2023 VQA datasets, and newly constructed examples created by experts based on internet search results. All samples are filtered for difficulty and quality to ensure reliable factual evaluation. The evaluation subset contains 1,013 English examples, reducing the influence of multilingual variation.
    \item \textbf{LiveVQA.} LiveVQA~\cite{fu2025livevqa} is a news-oriented VQA benchmark built from content collected from major international media outlets, such as CNN and BBC. It contains 3,602 image–question pairs spanning 14 categories, including science and sports. The questions are generated with GPT-4o and cover a broad range of difficulty levels, from simple visual recognition to more complex reasoning over accompanying textual context.
\end{itemize}

\subsection{Video Understanding Benchmarks}
To evaluate the generalization ability of VideoSearcher on general video understanding tasks, we primarily adopt the following benchmarks:
\begin{itemize}[leftmargin=*, itemsep=2pt, topsep=2pt, parsep=0pt]
    \item \textbf{MMVU.} MMVU~\cite{zhao2025mmvu} is an expert-level video understanding benchmark for knowledge-intensive reasoning over specialized-domain videos. It contains 3,000 expert-annotated QA examples from 1,529 videos, including 1,000 validation and 2,000 test samples. The dataset spans 27 subjects across Science, Healthcare, Humanities \& Social Sciences, and Engineering, and includes both multiple-choice and open-ended questions. MMVU evaluates whether models can integrate dynamic visual evidence with domain-specific knowledge for expert-level video reasoning.

    \item \textbf{TempCompass.} TempCompass~\cite{liu2024tempcompass} evaluates the temporal perception ability of Video LLMs. It contains 410 open-domain videos, 500 annotated meta-information items, and 7,540 task instructions. The benchmark covers five temporal aspects: Action, Speed, Direction, Attribute Change, and Event Order, with tasks in multiple-choice QA, yes/no QA, caption matching, and caption generation formats. By constructing videos with similar static content but different temporal dynamics, TempCompass tests whether models truly understand temporal changes across frames.

    \item \textbf{VideoMMMU.} Video-MMMU~\cite{hu2025video} is a professional video benchmark for evaluating knowledge acquisition from educational videos. It contains 300 college-level videos and 900 human-annotated questions across six disciplines: Art, Business, Science, Medicine, Humanities, and Engineering. The questions cover three cognitive stages: Perception, Comprehension, and Adaptation. Video-MMMU measures whether models can identify key information, understand introduced concepts, and transfer learned knowledge to related problems.

    \item \textbf{VideoMathQA.} VideoMathQA~\cite{rasheed2025videomathqa} evaluates mathematical reasoning over instructional videos. It contains 420 manually annotated video--question pairs across 10 mathematical domains, with videos ranging from 10 seconds to over one hour. The questions cover Problem Focused, Concept Transfer, and Deep Comprehension reasoning, requiring models to integrate visual content, temporal context, and mathematical knowledge. Each sample includes expert-annotated reasoning steps, supporting evaluation of both final answers and reasoning quality.
\end{itemize}

\section{Experimental Analysis}
\subsection{Tool Invocation Analysis}
\begin{figure}[!t]
    \centering
    \includegraphics[width=1\linewidth]{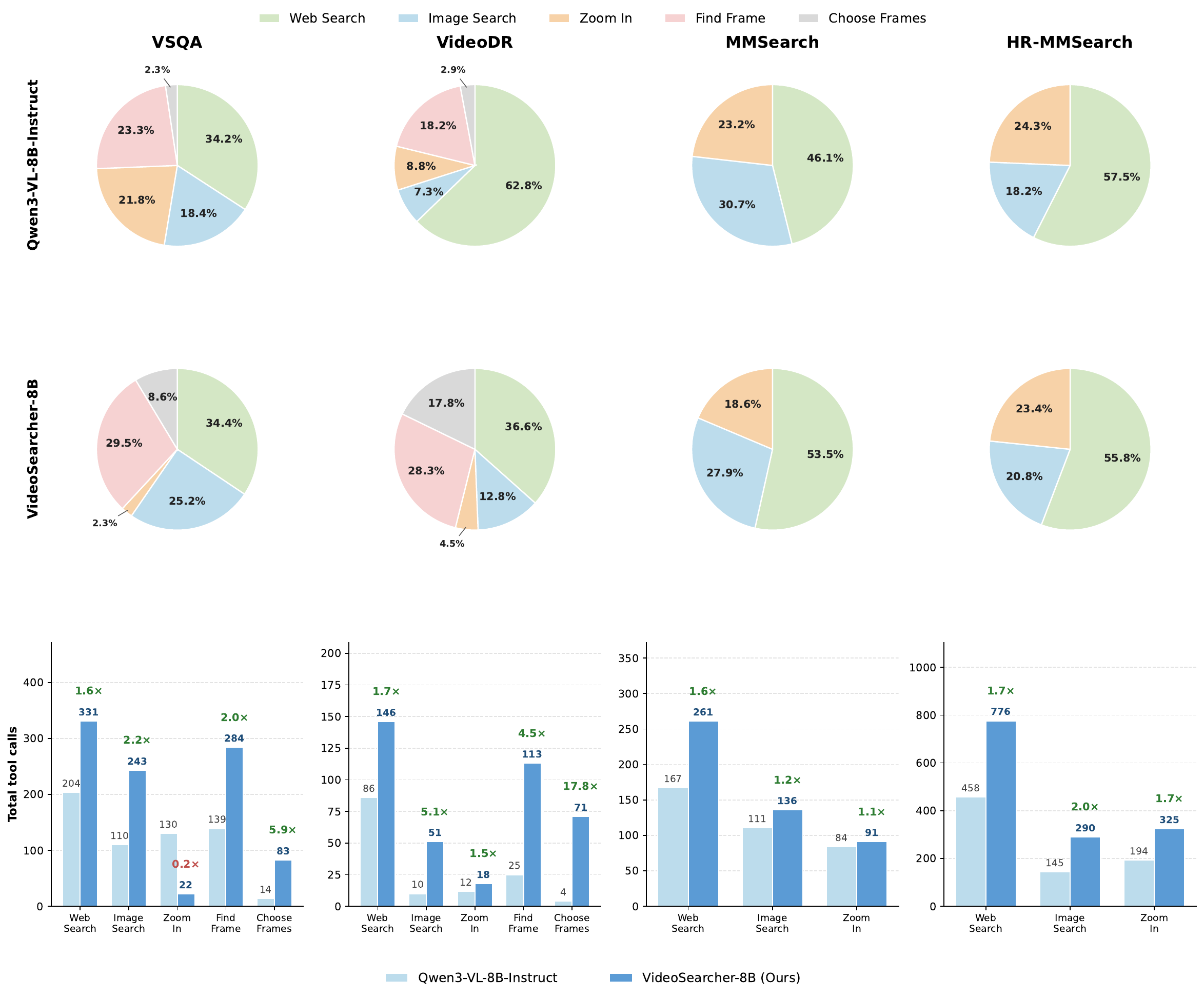}
    \caption{Comparison of tool usage patterns between Qwen3-VL-8B-Instruct and VideoSearcher-8B across video deep research and image search-oriented benchmarks.}
    \label{fig:tool_invocation}
\end{figure}

To better understand how VideoSearcher changes the model's tool-use behavior, we analyze the tool invocation patterns of Qwen3-VL-8B-Instruct and VideoSearcher-8B across four search-oriented benchmarks, as shown in Fig.~\ref{fig:tool_invocation}. Overall, VideoSearcher invokes tools more actively and more purposefully than the base model, indicating that our training encourages the model to acquire evidence through both video grounding and external retrieval.

\noindent \textbf{VDR Benchmarks.} On the two video-centric benchmarks, VSQA and VideoDR, VideoSearcher substantially increases the use of video localization tools. Compared with the base model, it invokes \texttt{find\_frame} and \texttt{choose\_frames} much more frequently, showing that the model learns to first ground the query in relevant video segments and key frames before searching for external evidence. Meanwhile, the numbers of \texttt{web\_search} and \texttt{image\_search} calls also increase notably, suggesting that VideoSearcher does not rely on video perception alone, but actively connects localized visual clues with open-web information. This behavior is consistent with the goal of Video Deep Research, where the model must jointly perform temporal localization, visual evidence grounding, and multimodal search.

\noindent \textbf{Zoom-in Behavior on VDR Benchmarks.}
We observe that \texttt{zoom\_in} is used relatively less frequently on the video benchmarks. A possible reason is that video inputs are typically represented by sampled frames with moderate visual resolution. As a result, once a relevant frame is localized, the model can often proceed directly to image search or web search without repeatedly magnifying small regions. 

\noindent \textbf{Image Search Benchmarks.} On MMSearch and HR-MMSearch, VideoSearcher also invokes search tools much more frequently than the base model.  This indicates that the learned tool-use policy transfers beyond video-centric tasks and improves general multimodal search behavior. In particular, on HR-MMSearch, which contains high-resolution visual inputs, VideoSearcher shows a clear increase in \texttt{zoom\_in} usage. This suggests that the model can adapt its tool strategy to the visual characteristics of the benchmark: when fine-grained details are important, it performs more local inspection before or alongside external search. 


\subsection{RL Analysis}
\begin{figure}[H]
    \centering
    \begin{subfigure}[t]{0.35\linewidth}
        \centering
        \includegraphics[width=\linewidth]{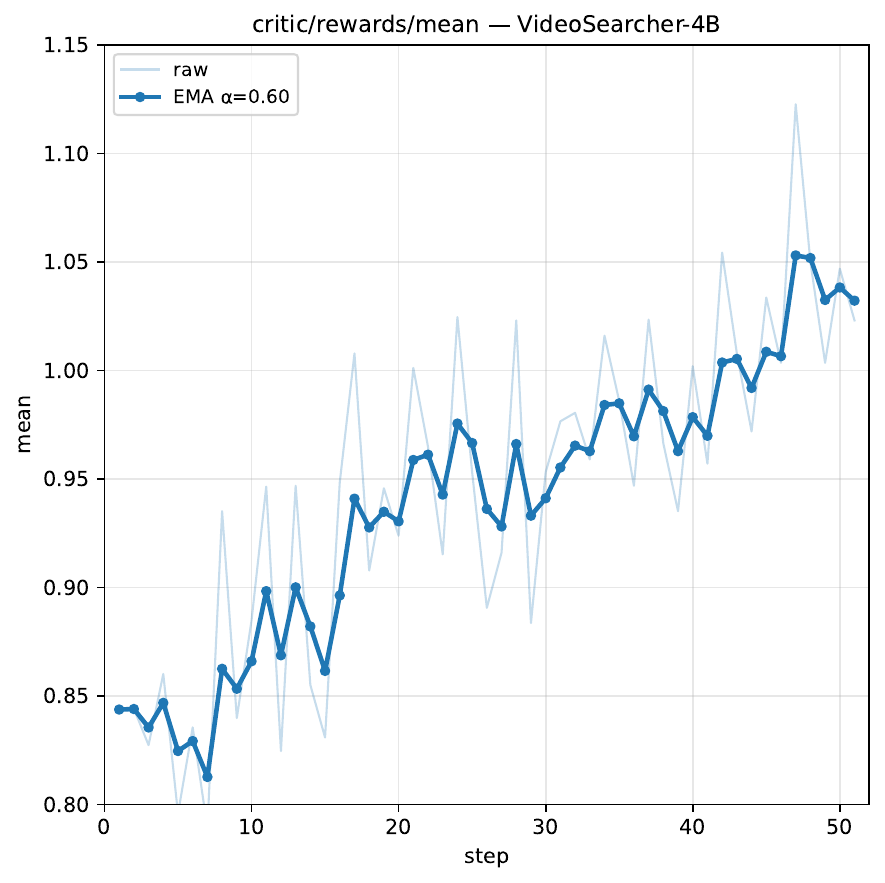}
        \caption{VideoSearcher-4B}
        \label{fig:rl_reward_4b}
    \end{subfigure}
    \hspace{0.06\linewidth}
    \begin{subfigure}[t]{0.35\linewidth}
        \centering
        \includegraphics[width=\linewidth]{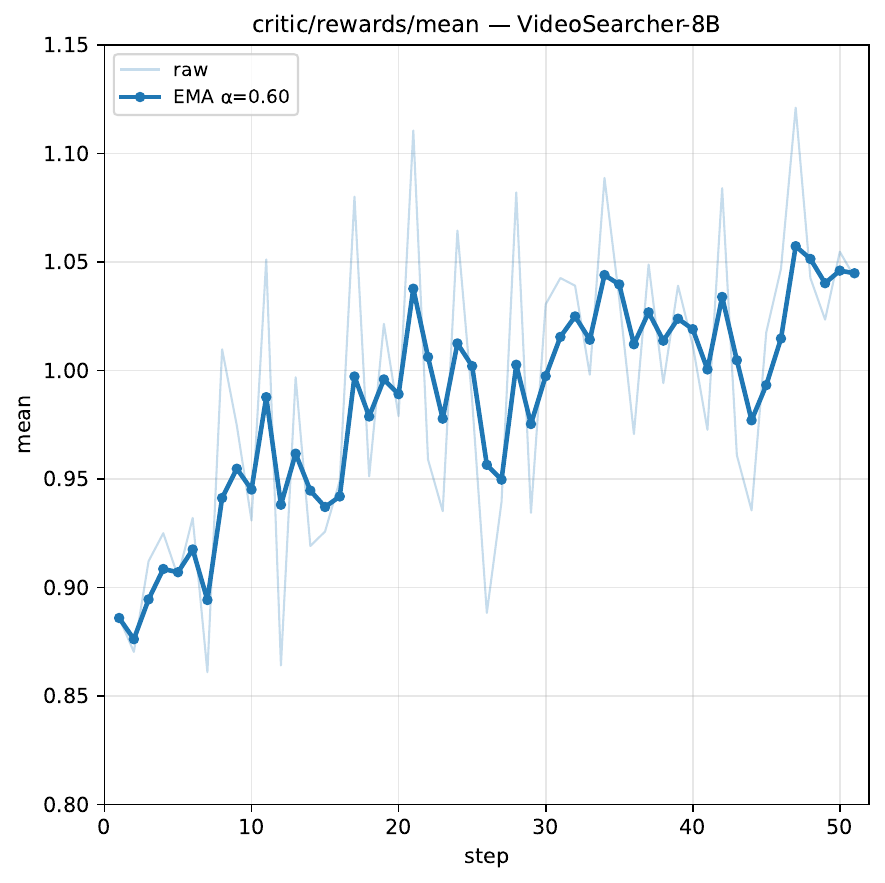}
        \caption{VideoSearcher-8B}
        \label{fig:rl_reward_8b}
    \end{subfigure}
    \caption{Training reward curves for VideoSearcher-4B and VideoSearcher-8B. The light curves show raw reward values, and the dark curves show EMA-smoothed trends.}
    \label{fig:rl_reward}
\end{figure}

Figure~\ref{fig:rl_reward} shows the training reward curves of VideoSearcher-4B and VideoSearcher-8B. For both model sizes, the mean reward increases steadily throughout RL training. VideoSearcher-4B improves from roughly 0.84 at the beginning of training to above 1.03 near the end, while VideoSearcher-8B rises from about 0.88 to around 1.05. The raw curves are noisy, which is expected in an online tool-interactive setting where rollouts depend on multi-turn decisions, external tool observations, and LLM-based judging. Nevertheless, the EMA curves show a consistent upward trend for both models.

The reward curves also suggest that the search-aware reward shaping does not prevent useful exploration. Although excessive or duplicate search calls are penalized, useful search behavior still increases when it benefits the task, as shown by the tool invocation statistics. This is important because simply rewarding every tool call can lead to over-searching, while only rewarding final answer correctness provides weak supervision for intermediate tool choices. By combining correctness-gated tool rewards with penalties for redundant search, BiSPO encourages the model to use tools when they help and to avoid unnecessary repeated retrieval.

\FloatBarrier

\section{Case Study}
\begin{figure}[p]
    \centering
    \includegraphics[height=0.78\textheight,keepaspectratio]{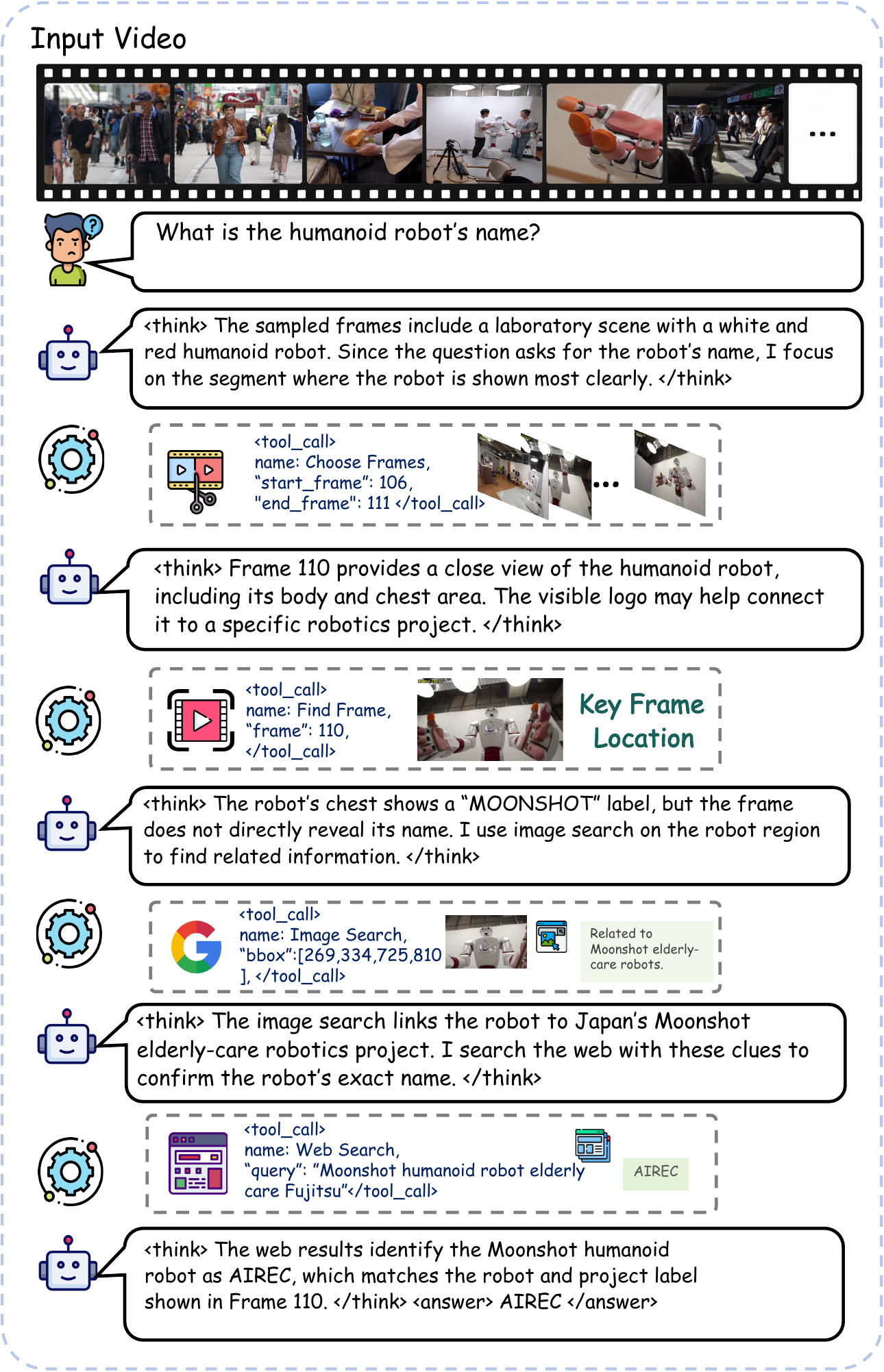}
    \caption{Successful case 1 of VideoSearcher.}
    \label{fig:case1}
\end{figure}

\begin{figure}[p]
    \centering
    \includegraphics[height=0.78\textheight,keepaspectratio]{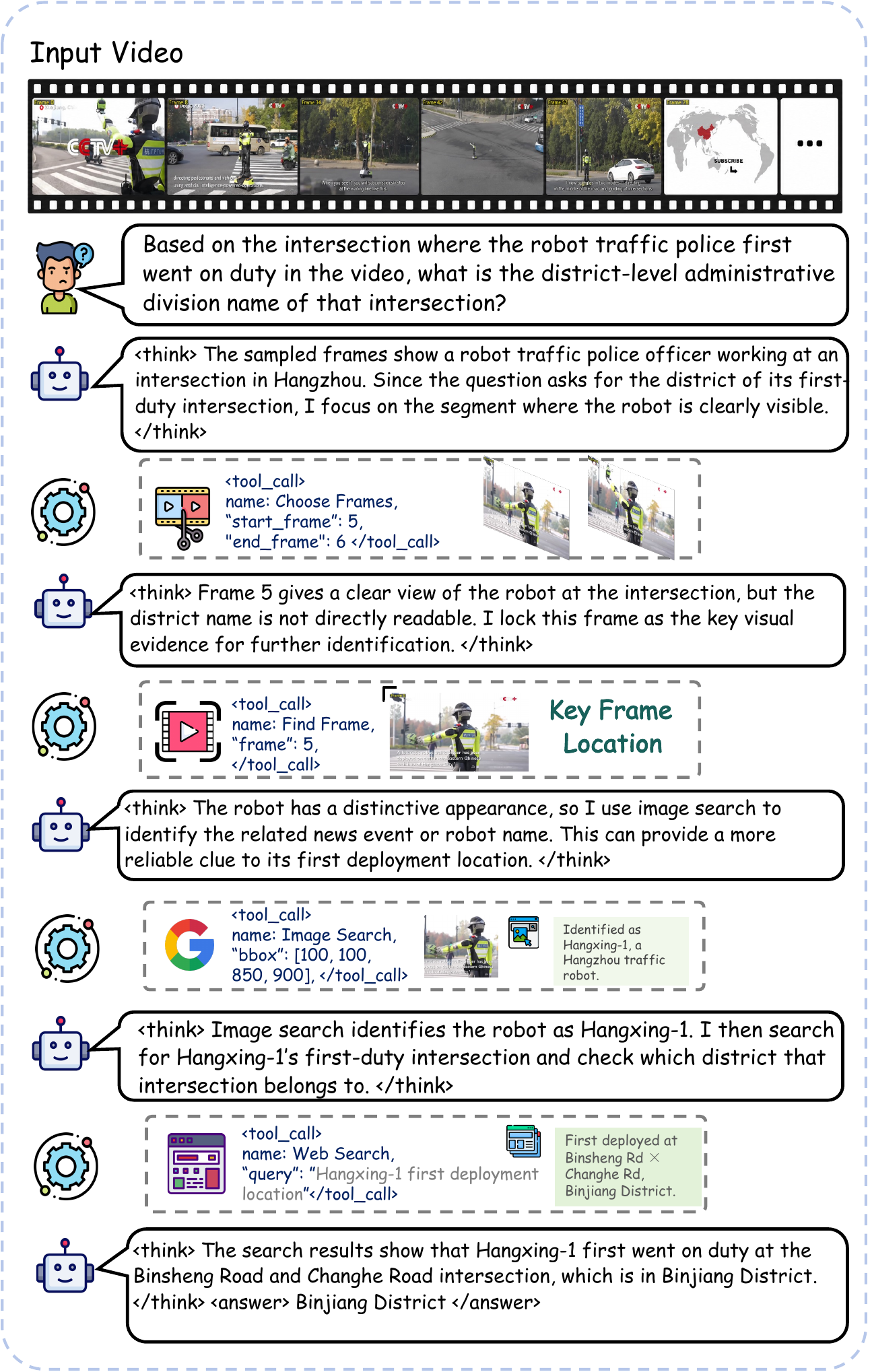}
    \caption{Successful case 2 of VideoSearcher.}
    \label{fig:case2}
\end{figure}

\begin{figure}[p]
    \centering
    \includegraphics[height=0.78\textheight,keepaspectratio]{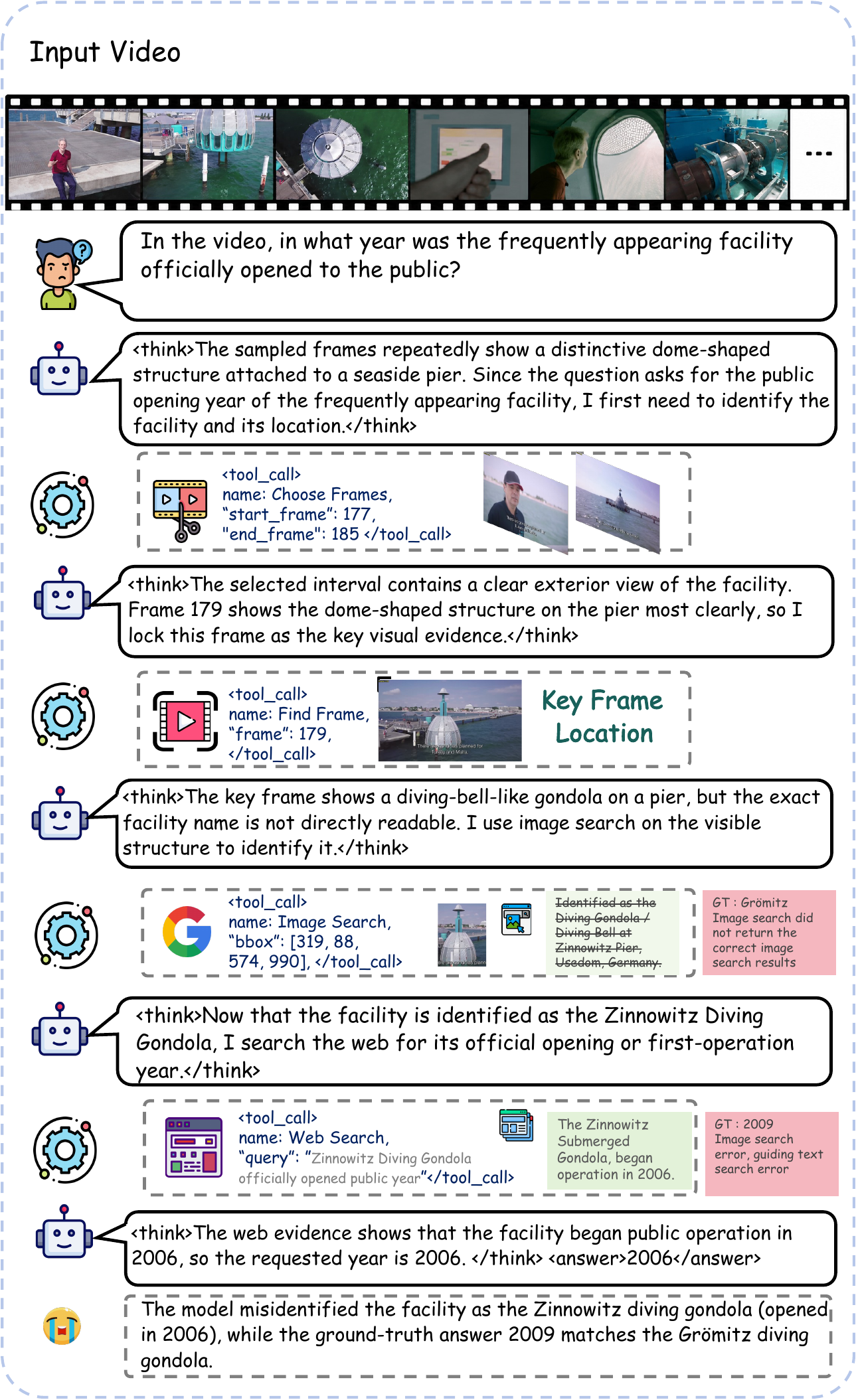}
    \caption{Failed case 3 of VideoSearcher.}
    \label{fig:case3}
\end{figure}

We further present qualitative case studies in Fig.~\ref{fig:case1}--Fig.~\ref{fig:case3} to analyze the reasoning behavior of VideoSearcher under successful and failed trajectories.

\noindent \textbf{Successful Cases.}  As shown in Fig.~\ref{fig:case1} and Fig.~\ref{fig:case2}, VideoSearcher can effectively decompose Video Deep Research problems into a sequence of temporally grounded and search-oriented actions. In the first case, the model first narrows the video to the relevant segment, locks the key frame containing the humanoid robot, and identifies the ``MOONSHOT'' visual cue on the robot body. Since the video itself does not directly reveal the robot name, VideoSearcher further invokes image search to associate the visual evidence with Japan's Moonshot elderly-care robotics project, and then uses web search to verify the exact entity, finally producing the correct answer \textit{AIREC}. Similarly, in the second case, the model localizes the frame where the robot traffic police is clearly visible, uses image search to identify it as \textit{Hangxing-1}, and then queries the web for its first deployment location. The retrieved evidence indicates that it first went on duty at the Binsheng Road and Changhe Road intersection, which belongs to Binjiang District. These cases demonstrate that VideoSearcher does not rely on a single retrieval step, but instead builds a coherent trajectory that connects video localization, visual entity recognition, external evidence retrieval, and final answer synthesis.

\noindent \textbf{Failed Cases.} Fig.~\ref{fig:case3} illustrates a representative failure case caused by erroneous retrieval feedback. The model correctly recognizes that the answer depends on identifying the dome-shaped facility repeatedly appearing in the video and follows a reasonable tool-use procedure: it selects the relevant temporal interval, locks a key frame, and performs image search on the visible structure. However, the image search tool incorrectly returns the facility as the Zinnowitz Diving Gondola, while the ground-truth visual entity is the Gr{\"o}mitz Diving Gondola. This incorrect retrieval result then propagates through the subsequent reasoning chain: the model formulates a web search query based on the wrong entity and retrieves the opening year of the Zinnowitz facility, leading to the incorrect answer \textit{2006} instead of the ground-truth answer \textit{2009}. This failure highlights a key limitation of open-world VDR agents: even when the model follows a valid and purposeful tool trajectory, its final answer can still be misled by noisy or incorrect search results. It suggests that future systems should incorporate stronger cross-source verification, retrieval uncertainty estimation, and visual-evidence consistency checking to reduce error propagation from external tools.

\FloatBarrier

\section{Prompt}
\label{app:prompt}
This section summarizes the prompts used in training and evaluation. We design different prompts for Video Deep Research, direct visual understanding, and tool-augmented image reasoning. All prompts follow a unified structured-output format to ensure consistent parsing, tool execution, and automatic evaluation.

\subsection{Training}
\noindent \textbf{VideoSearcher Training Prompt.}
As shown in Fig.~\ref{fig:videodeepresearch-training-prompt}, the training prompt defines the model as a Video Deep Research assistant that solves user queries by jointly grounding visual clues in videos and retrieving external knowledge. 
The prompt specifies the video representation, available tools, tool dependency rules, and the required \texttt{<think>}, \texttt{<tool\_call>}, and \texttt{<answer>} output format. 
This prompt is used to construct supervised trajectories and initialize the model with basic video navigation, frame localization, region inspection, multimodal search, and answer synthesis behaviors. 
Following the video-centric training pipeline, these trajectories provide the cold-start interaction protocol before online RL optimization.

\begin{figure*}[p]
\centering
\begin{tcolorbox}[
     title=\textbf{System Prompt for VideoSearcher Training},
     appendixprompt
]
\scriptsize

\noindent\textbf{Role.}
You are an advanced \textbf{Video Deep Research reasoning assistant}. Given a user query that requires combining visual clues from a video with external knowledge, your task is to solve the problem step-by-step by deeply navigating the video and searching the web.

\smallskip
\noindent\textbf{Video Context.}
The original video has been converted to \textbf{1 frame per second} (1 fps). The original frame index, such as \textit{Frame 1} or \textit{Frame 10}, is watermarked in the top-left corner of every frame. You are initially provided with uniformly sampled frames from the video.

\smallskip
\noindent\textbf{Available Tools.}
You may call one or more tools to assist with the user query. The available tools are summarized as follows:

\begin{enumerate}[leftmargin=*, noitemsep, topsep=2pt]
    \item \textbf{\texttt{choose\_frames}}: Select a specific time interval to investigate further. The system returns uniformly sampled frames from this interval. Use this when the answer lies within a specific video segment.
    \item \textbf{\texttt{find\_frame}}: Lock onto a specific single frame index for close examination. The system returns the exact frame.
    \item \textbf{\texttt{zoom\_in}}: Magnify a specific region of the currently locked frame for detailed visual analysis. The bounding box is represented as \texttt{[x1, y1, x2, y2]} in 0--1000 relative coordinates.
    \item \textbf{\texttt{image\_search}}: Reverse image search a specific region of the locked frame to identify unknown entities. Use this when the visual entity is not yet recognized.
    \item \textbf{\texttt{web\_search}}: Search the web using a text query. Use this when the entity is already recognized and external knowledge or factual verification is needed.
\end{enumerate}

\smallskip
\noindent\textbf{Tool Dependency and Workflow Rules.}
The reasoning process must follow the tool dependency rules below:

\begin{enumerate}[leftmargin=*, noitemsep, topsep=1pt]
    \item \textbf{Initial Action}: Directly call \texttt{find\_frame} if the target is obvious in the initially sampled frames. Otherwise, call \texttt{choose\_frames} to narrow down the relevant interval.
    \item \textbf{Interval Narrowing}: \texttt{choose\_frames} may be called multiple times in succession to progressively narrow the search interval.
    \item \textbf{Frame Locking}: Before using \texttt{zoom\_in}, \texttt{image\_search}, or \texttt{web\_search}, the assistant must first call \texttt{find\_frame} to lock onto a specific frame.
    \item \textbf{Reset Rule}: Each call to \texttt{choose\_frames} resets the current frame lock. After any \texttt{choose\_frames} call, the assistant must call \texttt{find\_frame} again before using detail/search tools or producing the final answer.
    \item \textbf{Detailing and Searching}: After a valid \texttt{find\_frame} call, the assistant may use \texttt{zoom\_in} for visual inspection, \texttt{image\_search} for unknown visual entities, or \texttt{web\_search} for external factual knowledge.
    \item \textbf{Ending Constraint}: The last tool call before the final answer must not be \texttt{choose\_frames}. If \texttt{choose\_frames} is called, the assistant must continue with either further interval narrowing or a \texttt{find\_frame} call before answering.
\end{enumerate}

\smallskip
\noindent\textbf{Output Format.}
At each turn, the assistant must either issue \textbf{one precise tool call} or provide the final answer. All outputs must begin with a thought process enclosed in \texttt{<think></think>} tags.

\noindent If reasoning continues, use the following format:
\begin{flushleft}
\ttfamily\scriptsize
<think> ... </think>\\
<tool\_call>\{"name": "<function-name>", "arguments": <args-json-object>\}</tool\_call>
\end{flushleft}

\noindent If the assistant is ready to conclude, use the following format:
\begin{flushleft}
\ttfamily\scriptsize
<think> ... </think>\\
<answer> Final answer to the user's query </answer>
\end{flushleft}

\end{tcolorbox}
\caption{System prompt for VideoSearcher reasoning with video navigation and external knowledge retrieval.}
\label{fig:videodeepresearch-training-prompt}
\end{figure*}

\subsection{Evaluation}
\noindent \textbf{VideoSearcher Inference Prompt.}
For agentic video evaluation, we use the inference prompt shown in Fig.~\ref{fig:videodeepresearch-inference-prompt}.  The prompt follows the same Video Deep Research protocol as the training prompt, but provides explicit function signatures and stricter inference-time constraints. The model is initially given 64 uniformly sampled frames and must decide whether to localize a temporal interval, lock a key frame, inspect a region, perform image search, or use web search. It also includes stop-searching rules to prevent unnecessary tool calls once sufficient evidence has been collected. This setting is used for evaluating VideoSearcher on VideoDR~\cite{liu2026watching} and our VideoSearch-QA benchmark.

\noindent \textbf{Direct Image Understanding Prompt.}
For direct image-based evaluation, we use the prompt in Fig.~\ref{fig:image-direct-prompt}.  The model answers directly from the provided image content without calling external tools.  This setting provides a tool-free baseline.

\noindent \textbf{Image Reasoning with Tools Prompt.}
For tool-augmented image reasoning baselines, we use the prompt shown in Fig.~\ref{fig:image-tool-prompt}. The model is allowed to invoke image zooming, text search, and reverse image search tools, following prior multimodal search and agentic reasoning settings~\cite{wu2025mmsearch, hong2025deepeyesv2}. At each turn, the model must either issue one valid tool call or provide the final answer. Unlike VideoSearcher, this prompt focuses on image-level reasoning and does not involve frame localization.

\noindent \textbf{Direct Video Understanding Prompt.}
For general video understanding benchmarks, we use the direct video prompt in Fig.~\ref{fig:video-direct-prompt}. The model answers only from the sampled video frames, supplemental images if provided, and its internal knowledge.  No tool calls or tool-like JSON outputs are allowed. This prompt is used to evaluate the model under the conventional video understanding setting, where the answer is expected to be inferred from the video itself rather than retrieved from external evidence.

\begin{figure*}[p]
\centering
\begin{tcolorbox}[
     title=\textbf{System Prompt for Model Inference},
     appendixprompt
]
\scriptsize

\noindent\textbf{Role.}
You are an advanced \textbf{Video Deep Research reasoning assistant}. Given a user query that requires combining visual clues from a video with external knowledge, your task is to solve the problem step-by-step by deeply navigating the video and searching the web.

\noindent\textbf{Video Context.}
The original video has been converted to \textbf{1 frame per second} (1 fps). The original frame index, such as \textit{Frame 1} or \textit{Frame 10}, is watermarked in the top-left corner of every frame. The assistant is initially provided with \textbf{64 uniformly sampled frames} from the video.

\noindent\textbf{Available Tools.}
The assistant may call one or more functions to solve the query. The available tools are provided as function signatures within \texttt{<tools></tools>} XML tags:

\begin{enumerate}[leftmargin=*, noitemsep, topsep=2pt]
    \item \textbf{\texttt{choose\_frames(start\_frame\_index, end\_frame\_index)}}: Select a specific time interval for further investigation. The system returns uniformly sampled frames from the selected interval. This tool is used when the answer is likely located within a particular video segment.
    
    \item \textbf{\texttt{find\_frame(frame\_index)}}: Lock onto a specific single frame for close examination. The system returns the exact frame corresponding to the specified frame index.
    
    \item \textbf{\texttt{zoom\_in(bbox)}}: Magnify a specific region of the currently locked frame for detailed visual analysis. The bounding box is represented as \texttt{[x1, y1, x2, y2]} in 0--1000 relative coordinates, where \texttt{0 <= x1 < x2 <= 1000} and \texttt{0 <= y1 < y2 <= 1000}.
    
    \item \textbf{\texttt{image\_search(bbox)}}: Perform reverse image search over a specific region of the locked frame to identify unknown visual entities. This tool is used when the assistant visually locates an entity but does not know its name or identity.
    
    \item \textbf{\texttt{web\_search(query)}}: Search the web using a text query. This tool is used when the assistant already recognizes the entity and needs external knowledge or factual verification.
\end{enumerate}

\noindent\textbf{Tool Dependency and Workflow Rules.}
The assistant must strictly follow the tool dependency rules below:

\begin{enumerate}[leftmargin=*, noitemsep, topsep=1pt]
    \item \textbf{Initial Action}: The assistant may directly call \texttt{find\_frame} if the target is obvious in the initial 64 sparse frames. Otherwise, it should call \texttt{choose\_frames} to narrow down the search interval.
    
    \item \textbf{Interval to Frame}: After calling \texttt{choose\_frames} and receiving the sub-frames, the assistant must call \texttt{find\_frame} to lock onto a specific single frame before performing any detailed action.
    
    \item \textbf{Detailing and Searching}: \texttt{zoom\_in} and \texttt{image\_search} must only be used after a successful \texttt{find\_frame} call. After locking a frame, the assistant may call \texttt{zoom\_in} to inspect details, then decide whether to use \texttt{image\_search} or \texttt{web\_search}. If the locked frame is already clear enough, the assistant may skip \texttt{zoom\_in} and directly call \texttt{image\_search} or \texttt{web\_search}.
    
    \item \textbf{Search Strategy Selection}: If an entity is visually located but not recognized, the assistant should use \texttt{image\_search(bbox)}. If the entity is recognized, the assistant should bypass image search and directly use \texttt{web\_search(query)}.
    
    \item \textbf{Retry Mechanism}: If the current search fails, yields incorrect information, or does not solve the query, the assistant may loop back to \texttt{choose\_frames} or \texttt{find\_frame} to explore another segment or frame. The assistant has a maximum of \textbf{10 attempts} to find the answer. If all attempts are exhausted, it should provide the best supported answer with an explicit note about remaining uncertainty.
\end{enumerate}

\noindent\textbf{Output Format.}
At each turn, the assistant must either issue \textbf{one precise tool call} or provide the final answer. All outputs must begin with a thought process enclosed in \texttt{<think></think>} tags.

\noindent If reasoning continues, the assistant must use:
\begin{flushleft}
\ttfamily\scriptsize
<think> ... </think>\\
<tool\_call>\{"name": "<function-name>", "arguments": <args-json-object>\}</tool\_call>
\end{flushleft}

\noindent If ready to conclude, the assistant must use:
\begin{flushleft}
\ttfamily\scriptsize
<think> ... </think>\\
<answer> Final answer to the user's query </answer>
\end{flushleft}

\noindent\textbf{Stop Searching Rules.}
The assistant must stop searching once sufficient evidence has been gathered:

\begin{enumerate}[leftmargin=*, noitemsep, topsep=1pt]
    \item If a search result already contains the requested fact or enough evidence to infer the answer, the assistant should stop using tools and provide the final \texttt{<answer>}.
    
    \item The assistant should not repeatedly refine the same web search query after it has already returned the same fact. Near-duplicate queries are not considered useful additional evidence.
    
    \item Near the final attempt, the assistant must use the evidence already collected and answer, rather than calling another tool only to confirm the same fact again.
    
    \item If the evidence is incomplete but the attempt budget is nearly exhausted, the assistant should give the best supported answer in \texttt{<answer>} and mention uncertainty only inside \texttt{<think>}.
\end{enumerate}

\end{tcolorbox}
\caption{System prompt for VideoSearcher reasoning with video navigation, external knowledge retrieval, and strict tool-use constraints.}
\label{fig:videodeepresearch-inference-prompt}
\end{figure*}

\begin{figure*}[p]
\centering
\begin{tcolorbox}[
     title=\textbf{System Prompt for Direct Image Understanding},
     appendixprompt
]
\scriptsize

\noindent\textbf{Role.}
You are an advanced \textbf{general image understanding assistant}. Given a user query about one or more images, your task is to answer directly from the provided visual content and any supplemental images.

\noindent\textbf{Image Context.}
The assistant is provided with one or more input images. When multiple images are provided, they may be referenced as \texttt{\textless image 1\textgreater}, \texttt{\textless image 2\textgreater}, etc. The assistant should carefully inspect the visible content, spatial relationships, objects, text, attributes, and scene context in the provided images.

\noindent\textbf{Direct Answer Rules.}
The assistant must answer under the following constraints:

\begin{enumerate}[leftmargin=*, noitemsep, topsep=1pt]
    \item \textbf{No Tool Use}: The assistant must not call tools, output tool-call tags, or produce tool-like JSON.
    
    \item \textbf{Evidence Scope}: The assistant must use only the provided images, supplemental visual inputs, and its internal knowledge.
    
    \item \textbf{Visual Grounding}: The assistant should base its answer on visible evidence in the image, including objects, regions, colors, text, spatial layout, actions, and relationships.
    
    \item \textbf{Multiple-image Reasoning}: If multiple images are provided, the assistant should compare and integrate evidence across the referenced images when necessary.
    
    \item \textbf{Multiple-choice Questions}: For multiple-choice questions, the final answer must contain only the option letter.
    
    \item \textbf{Short-answer Questions}: For short-answer questions, the final answer must contain only a concise answer.
    
    \item \textbf{Incomplete Evidence}: If the visual evidence is incomplete or ambiguous, the assistant should answer with the best supported conclusion.
\end{enumerate}

\noindent\textbf{Output Format.}
The assistant may include brief reasoning in \texttt{\textless think\textgreater\textless/think\textgreater}, followed by the final answer in \texttt{\textless answer\textgreater\textless/answer\textgreater}. The assistant must not output previous reasoning chains. The required format is:

\begin{flushleft}
\ttfamily\scriptsize
<think> ... </think>\\
<answer> Final answer to the user's query </answer>
\end{flushleft}

\end{tcolorbox}
\caption{System prompt for direct image understanding without tool use.}
\label{fig:image-direct-prompt}
\end{figure*}
\begin{figure*}[p]
\centering
\begin{tcolorbox}[
     title=\textbf{System Prompt for Image Reasoning with Tools},
     appendixprompt
]
\scriptsize

\noindent\textbf{Role.}
You are a \textbf{step-by-step reasoning assistant}. Given a question, your task is to solve the problem one substep at a time.

\noindent\textbf{Guiding Principles.}
At each turn, the assistant must take exactly one of the following actions:

\begin{enumerate}[leftmargin=*, noitemsep, topsep=1pt]
    \item Issue one specific tool call enclosed in \texttt{\textless tool\_call\textgreater\textless/tool\_call\textgreater} tags.
    \item Provide the final answer enclosed in \texttt{\textless answer\textgreater\textless/answer\textgreater} tags.
\end{enumerate}

All outputs must begin with a reasoning step enclosed in \texttt{\textless think\textgreater\textless/think\textgreater} tags, explaining the current reasoning state and the next action. The assistant must not output previous reasoning chains.

\noindent\textbf{Output Format.}
The assistant must strictly follow one of the two formats below.

\noindent If reasoning continues, use:
\begin{flushleft}
\ttfamily\scriptsize
<think> Current reasoning and next plan </think>\\
<tool\_call> One precise tool call </tool\_call>
\end{flushleft}

\noindent If ready to conclude, use:
\begin{flushleft}
\ttfamily\scriptsize
<think> Summarize the reasoning and derive the answer </think>\\
<answer> Final answer </answer>
\end{flushleft}

\noindent\textbf{Available Tools.}
The assistant may call one or more functions to assist with the user query. The available tools are provided as function signatures within \texttt{\textless tools\textgreater\textless/tools\textgreater} XML tags:

\begin{enumerate}[leftmargin=*, noitemsep, topsep=1pt]
    \item \textbf{\texttt{image\_zoom\_in\_tool(bbox\_2d, label, img\_idx)}}: Zoom in on a specific region of an image by cropping it according to a bounding box and an optional object label. The bounding box is represented as \texttt{[x1, y1, x2, y2]}, where \texttt{(x1, y1)} is the top-left corner and \texttt{(x2, y2)} is the bottom-right corner. The parameter \texttt{img\_idx} specifies the index of the image to inspect, starting from 0.

    \item \textbf{\texttt{text\_search\_tool(query)}}: Search the web for text information related to the query. This tool is used when factual, news, or external information is required.

    \item \textbf{\texttt{image\_search\_tool()}}: Perform a reverse image search to find similar images and related information. This tool can help identify objects, places, or provide additional context about the image.
\end{enumerate}

\noindent\textbf{Tool Call Format.}
For each function call, the assistant must return a JSON object with the function name and arguments enclosed within \texttt{\textless tool\_call\textgreater\textless/tool\_call\textgreater} XML tags:

\begin{flushleft}
\ttfamily\scriptsize
<tool\_call>\\
\{"name": "<function-name>", "arguments": <args-json-object>\}\\
</tool\_call>
\end{flushleft}

\end{tcolorbox}
\caption{System prompt for step-by-step image reasoning with zoom, text search, and reverse image search tools.}
\label{fig:image-tool-prompt}
\end{figure*}

\begin{figure*}[p]
\centering
\begin{tcolorbox}[
     title=\textbf{System Prompt for Direct Video Understanding},
     appendixprompt
]
\scriptsize

\noindent\textbf{Role.}
You are an advanced \textbf{general video understanding assistant}. Given a user query about a video, your task is to answer directly from the provided video frames and any supplemental images.

\noindent\textbf{Video Context.}
The input video has already been converted to \textbf{1 frame per second} (1 fps). The assistant is provided with uniformly sampled frames from the video. Some questions may include supplemental images referenced as \texttt{\textless image 1\textgreater}, \texttt{\textless image 2\textgreater}, etc.; these images are provided after the sampled video frames.

\noindent\textbf{Direct Answer Rules.}
The assistant must answer under the following constraints:

\begin{enumerate}[leftmargin=*, noitemsep, topsep=1pt]
    \item \textbf{No Tool Use}: The assistant must not call tools, output tool-call tags, or produce tool-like JSON.
    
    \item \textbf{Evidence Scope}: The assistant must use only the provided video frames, supplemental images, and its internal knowledge.
    
    \item \textbf{Multiple-choice Questions}: For multiple-choice questions, the final answer must contain only the option letter.
    
    \item \textbf{Short-answer Questions}: For short-answer questions, the final answer must contain only a concise answer.
    
    \item \textbf{Incomplete Evidence}: If the visual evidence is incomplete, the assistant should answer with the best supported conclusion.
\end{enumerate}

\noindent\textbf{Output Format.}
The assistant may include brief reasoning in \texttt{\textless think\textgreater\textless/think\textgreater}, followed by the final answer in \texttt{\textless answer\textgreater\textless/answer\textgreater}. The required format is:

\begin{flushleft}
\ttfamily\scriptsize
<think> ... </think>\\
<answer> Final answer to the user's query </answer>
\end{flushleft}

\end{tcolorbox}
\caption{System prompt for direct video understanding without tool use.}
\label{fig:video-direct-prompt}
\end{figure*}
\FloatBarrier

\section{Moral and Ethical Statement}

\paragraph{Data Source and Usage Compliance.}
All videos and web materials used in our data construction are collected from publicly accessible online sources. We download and process the data in compliance with the corresponding platform policies and use them solely for academic research and evaluation. The collected videos are used only to construct video-grounded QA instances and tool-interaction trajectories. We will release our dataset and models to the community.

\end{document}